\documentclass[sn-mathphys-num]{sn-jnl}

\usepackage{graphicx}%
\usepackage{multirow}%
\usepackage{amssymb,amsfonts}%
\usepackage{amsthm}%
\usepackage{mathrsfs}%
\usepackage[title]{appendix}%
\usepackage{xcolor}%
\usepackage{textcomp}%
\usepackage{manyfoot}%
\usepackage{booktabs}%
\usepackage{algorithm}%
\usepackage{algorithmicx}%
\usepackage{algpseudocode}%
\usepackage{listings}%
\usepackage{tabularx,ragged2e}

\newcolumntype{C}{>{\Centering\arraybackslash}X} 
\usepackage[cmex10]{amsmath}
\usepackage{tikz}
\newcommand{\xmark}{%
\tikz[scale=0.2] {
    \draw[line width=0.7,line cap=round] (0,0) to [bend left=6] (1,1);
    \draw[line width=0.7,line cap=round] (0.2,0.95) to [bend right=3] (0.8,0.05);
}}
\newcommand{\cmark}{%
\tikz[scale=0.2] {
    \draw[line width=0.7,line cap=round] (0.25,0) to [bend left=10] (1,1);
    \draw[line width=0.8,line cap=round] (0,0.35) to [bend right=1] (0.23,0);
}}
\usepackage{bm}
\usepackage{threeparttable}
\usepackage{nicefrac}
\DeclareMathOperator{\E}{\mathbb{E}}
\usepackage{algorithm}
\usepackage{algpseudocode}
\algnewcommand\algorithmicforeach{\textbf{for each}}
\algdef{S}[FOR]{ForEach}[1]{\algorithmicforeach\ #1\ \algorithmicdo}

\theoremstyle{thmstyleone}%
\theoremstyle{thmstyletwo}%

\theoremstyle{thmstylethree}%

\begin{document}

\title[DG-ML]{Domain Generalization through Meta-Learning: A Survey}

\author*[1]{\fnm{Arsham} \sur{Gholamzadeh Khoee}}\email{khoee@chalmers.se}

\author[1]{\fnm{Yinan} \sur{Yu}}\email{yinan@chalmers.se}

\author[1]{\fnm{Robert} \sur{Feldt}}\email{robert.feldt@chalmers.se}

\affil*[1]{\orgdiv{Department of Computer Science and Engineering}, \orgname{Chalmers University of Technology}, \orgaddress{ \city{Gothenburg}, \country{Sweden}}}


\abstract{Deep neural networks (DNNs) have revolutionized artificial intelligence but often lack performance when faced with out-of-distribution (OOD) data, a common scenario due to the inevitable domain shifts in real-world applications. This limitation stems from the common assumption that training and testing data share the same distribution—an assumption frequently violated in practice. Despite their effectiveness with large amounts of data and computational power, DNNs struggle with distributional shifts and limited labeled data, leading to overfitting and poor generalization across various tasks and domains.
Meta-learning presents a promising approach by employing algorithms that acquire transferable knowledge across various tasks for fast adaptation, eliminating the need to learn each task from scratch.
This survey paper delves into the realm of meta-learning with a focus on its contribution to domain generalization. We first clarify the concept of meta-learning for domain generalization and introduce a novel taxonomy based on the feature extraction strategy and the classifier learning methodology, offering a granular view of methodologies. 
Additionally, we present a decision graph to assist readers in navigating the taxonomy based on data availability and domain shifts, enabling them to select and develop a proper model tailored to their specific problem requirements.
Through an exhaustive review of existing methods and underlying theories, we map out the fundamentals of the field. Our survey provides practical insights and an informed discussion on promising research directions.}




\maketitle
\vspace{7pt}
\section{Introduction}
\label{sec:intro}

Traditional machine learning approaches assume training and testing data are independent and identically distributed (i.i.d.). However, this assumption often fails in practice due to variations in data acquisition conditions, such as changes in sensor types, lighting conditions, or environmental factors, leading to substantial performance degradation when models encounter new domains~\cite{khosla2012undoing}; thus, they cannot cope with the out-of-distribution (OOD) challenge~\cite{ganin2016domain}. A domain can be defined as a specific distribution of data characterized by its own set of features, statistical properties, and underlying concepts~\cite{sugiyama2006mixture}. Domain generalization (DG) and domain adaptation (DA) techniques evolved to deal with domain shifts to address out-of-distribution problems~\cite{ghifary2015domain, tzeng2014deep}. When we have two distinct domains, namely the source and target domains, we can apply domain adaptation techniques, which focus on adapting a model trained on a source domain to perform well on a target domain. Meanwhile, domain generalization aims to generalize over any domain, allowing models trained on one set of domains to perform well on unseen domains.
It is important to note that DA techniques assume access to target domain data~\cite{patel2015visual, long2016unsupervised}, whereas DG techniques assume no access to the target domain data, which makes it a much harder problem to solve. Compared to DA, DG is more applicable in practice. First, machine learning models are trained on a specific large dataset to perform well on that exact task. However, in real-world scenarios, we often face new tasks with limited labeled data available. Essentially, collecting labeled data and retraining the model for such "unseen" domains can be extremely costly and time-consuming. Also, it is often impossible to enumerate all the "unseen" domains in advance. Accordingly, enhancing the generalization capability of machine learning models is crucial.

Meta-learning has emerged to enable a model to quickly adapt and learn from a small amount of data or even generalize to unseen tasks~\cite{huisman2021survey}. Meta-learning algorithms leverage prior knowledge, patterns, and experiences acquired from similar or related tasks to make learning transferable~\cite{ravi2016optimization}. Recently, some meta-learning algorithms have been used for domain generalization, allowing models to generalize across various domains with limited or no access to all domains during training. These algorithms learn a more generalized representation by identifying patterns that generalize across unseen domains. For instance, consider the PACS dataset~\cite{li2017deeper}, which comprises images from four distinct domains: photo, art, cartoon, and sketch. The goal is to train a model on a subset of these domains, such as photos, art, and cartoons, in such a way that it can also accurately classify sketches despite never having ``seen" sketches during training. This challenges the model to extract domain-invariant patterns essential for classifying images across all four domains. This ``learning-to-generalize" paradigm aligns with the human learning process, where knowledge acquired from multiple situations enables individuals to adapt quickly to new situations~\cite{finn2017model}.

Recently, there has been a growing interest in developing meta-learning methods for domain generalization. These methods aim to address the limitations of conventional machine learning approaches by incorporating meta-level learning procedures that enhance a model's ability to generalize across domains. More specifically, these models achieve cross-domain generalization by learning from a variety of tasks and domains during training, which prepares them to handle distributional shifts and enables more effective generalization. It is important to note that in domain generalization via meta-learning, the assumption is that there are zero target domain examples during training, contrasting with typical meta-learning that often focuses on few-shot learning for task adaptation.

Wang et al.~\cite{wang2022generalizing} and Zhou et al.~\cite{zhou2022domain} discussed recent advances in domain generalization, covering the formal definition of the problem, related fields, and theoretical foundations, and they highlight commonly used datasets and applications. Additionally, Liu et al.~\cite{liu2021towards} recently explored the OOD generalization problem, which addresses scenarios where the test data distribution differs from the training data. Their review encompasses methodologies categorized into unsupervised representation learning, supervised model learning, and optimization.
Viltala et al.~\cite{vilalta2002perspective} published one of the first surveys on meta-learning to formalize self-adaptive learner models.
Huisman et al.~\cite{huisman2021survey} later provided a comprehensive survey on meta-learning to explore the common challenge of how to leverage meta-knowledge to improve the performance of learning algorithms.
Hospedales et al.~\cite{hospedales2021meta} also discussed various perspectives on meta-learning by providing a novel taxonomy and investigating the utility of the most commonly used meta-learning algorithms. Vettoruzzo et al.~\cite{vettoruzzo2024advances} have recently examined meta-learning algorithms, discussing their benefits and challenges, and categorizing methods into metric-, model-, and optimization-based techniques, with a focus on achieving consistent evaluation metrics and computational efficiency. 
In contrast to previous works, to our best knowledge, this paper is the first survey collecting literature and highlighting the overall potential of meta-learning for domain generalization. In this paper, we offer a fresh perspective and a new taxonomy for classifying meta-learning approaches designed for generalization to unseen domains.

As a more detailed explanation, this survey paper \textit{provides a systematic overview of existing meta-learning methods tailored for domain generalization}. We introduce the concept of domain generalization and clarify its significance in real-world machine learning applications. Subsequently, we delve into the theoretical foundations of meta-learning and explain its applicability to the domain generalization setting. The paper reviews various frameworks and methodologies employed in existing meta-learning approaches for domain generalization, highlighting their strengths and limitations. Additionally, we present a taxonomy to help with the effective categorization of these algorithms. Furthermore, we survey the datasets and evaluation protocols prevalent in this research area and point out the principal challenges and open questions for future study. Our goal is to present an extensive reference for researchers and practitioners interested in understanding and advancing the field of meta-learning for domain generalization.

The rest of this survey is structured as follows. In Section~\ref{sec:backgeound}, we provide background information on the topic and introduce the fundamental concepts of meta-learning and its applicability to the domain generalization setting. Section~\ref{sec:taxonomy} presents a taxonomy that can be used to categorize techniques that will be discussed in this survey. We then proceed to explore the different frameworks of meta-learning methods for domain generalization in Section~\ref{sec:methodologies}. Section~\ref{sec:evaluation} discusses the widely adopted datasets and evaluation protocols employed in the field. In Section~\ref{sec:applications}, we cover the crucial applications of meta-learning for domain generalization and its significance in practical machine learning scenarios. Section~\ref{sec:discussion} discusses the significance of the findings, key challenges, and promising research directions. Finally, Section~\ref{sec:conclusion} summarizes the main points of the survey and their implications.

\vspace{7pt}
\section{Background}
\label{sec:backgeound}
While deep learning models have achieved remarkable results in specific tasks, developing models that can generalize across tasks and domains remains a challenge. Initially, we will provide a concise overview of related research areas, highlighting their unique characteristics, with a comparison of the discussed learning paradigms available in Table~\ref{table:learning}. Subsequently, we will delve into more details of meta-learning and domain generalization, along with providing a formalization of meta-learning for domain generalization, as it represents a research area specifically aimed at addressing the challenges of generalization across various tasks and domains.

\subsection{Different Learning Paradigms}
\label{subsec:learning}
In real-world applications, data availability is often limited and conditions change over time. A practical machine learning model must adapt to these changes without retraining from scratch, as traditional retraining is resource-intensive and inefficient. To address this challenge, several learning paradigms have been introduced. This section briefly introduces the key learning paradigms to build intuition for designing efficient machine learning models.

\subsubsection{Incremental Learning}
Incremental learning involves gradually training a model on new data over time. The model is updated incrementally as new data arrives, often in batches. However, previous data is often not retained, so incremental learning can suffer from catastrophic forgetting of earlier concepts. It is also necessary to carefully design training data to avoid bias and ensure that the AI system can generalize from the new data~\cite{wu2019large, he2011incremental,castro2018end}. The other important challenge is to deal with non-stationarity in the data, which can prevent the AI system from converging on a solution. Incremental learning can be done online or offline~\cite{he2020incremental}.

\subsubsection{Online Learning}
Online learning involves training a model on data that becomes available in a sequential order, sample-by-sample or mini-batch by mini-batch~\cite{hoi2021online}. The model is updated each time a new sample/mini-batch arrives. Online learning algorithms aim to perform updates efficiently to accommodate new data but do not explicitly retain old knowledge.

\subsubsection{Continual Learning}
Continual learning refers to the ability of a model to learn sequentially over time from a stream of data. The goal is to learn new tasks without forgetting previous knowledge, handling different distributions in the data over time~\cite{de2019continual}. It involves retaining previously learned knowledge and effectively integrating new knowledge.

\subsubsection{Transfer Learning}
Transfer learning refers to the process of reusing a pre-trained model as the basis for a new model, where a model trained on one task is repurposed for a second related task to allow rapid progress when modeling the second task~\cite{pan2009survey}. Essentially, transfer learning involves using the knowledge learned from one task to improve the performance of another related task.

\subsubsection{Multi-task Learning}
Multi-task learning is a machine learning approach in which we try to learn multiple tasks simultaneously, optimizing multiple loss functions at once. The goal of multi-task learning is to improve learning efficiency and prediction accuracy for the task-specific models, when compared to training the models separately~\cite{yang2014unified}. It leverages shared representations, which can help other tasks be learned better.

\subsubsection{Meta-Learning}
Meta-learning refers to the process of learning how to learn~\cite{finn2017model}. It involves training a model to learn new tasks quickly with minimal data by leveraging knowledge learned from previous tasks. The goal of meta-learning is to learn a good initialization of the model's parameters that can be adapted quickly to new tasks with only a few samples (also referred to as few-shot learning)~\cite{snell2017prototypical}. In other words, it transfers knowledge from many source tasks, so it can learn a new task more quickly, more proficiently, and more stably~\cite{hospedales2021meta}.

\subsubsection{Domain Adaptation}
DA focuses on adapting a model from one source domain to perform well on a related target domain~\cite{wang2018deep, csurka2017comprehensive, bousmalis2016domain,long2015learning, ben2006analysis}. It assumes some similarity or overlap between the two domains. It leverages labeled data from the source domain and unlabeled/limited labeled data from the target domain. Also, it is closely related to transfer learning, which also aims to transfer knowledge from the source domain to the target domain.

\subsubsection{Domain Generalization}
DG goes a step further in comparison to DA. It involves training a model on multiple source domains with the goal of making it perform well on any unseen or new target domains~\cite{muandet2013domain,li2017deeper,shankar2018generalizing}. So, it does not assume access to target domain data during training. DG addresses the challenge of adapting to diverse and unknown domains by learning representations that capture the underlying invariant factors across various domains.

\begin{table*}[ht]
\renewcommand{\arraystretch}{1.3}
\caption{Comparison of Different Learning Paradigms Highlighting Similarities and Dissimilarities Among Meta-Learning, Domain Generalization, and Related Learning Approaches.}
\label{table:learning}
\centering
\resizebox{1\textwidth}{!}{
\begin{threeparttable}
\begin{tabular}{c c c c c}
\hline
Learning Paradigm & $p(X)$ Consistency & $Y$ Consistency & Test Access & $\hat{p}(Y | X; \theta\tnote{1}~)$ \\ 
\hline
\hline
Incremental/Online Learning \cite{hoi2021online}  & $\cmark$ & $\cmark$ & $\cmark$ &$\min_{\theta} \E_{ p(\mathcal{D})} [\ell(\mathcal{D}; \theta)]$ \\
Continual Learning \cite{de2021continual}  & $\xmark$ & $\cmark$ & $\cmark$ & $\min_{\theta} \E_{p(\mathcal{T})} \E_{p(\mathcal{D}_\tau)} [\ell(\mathcal{D}_\tau; \theta)]$ \\
Transfer Learning \cite{tripuraneni2020theory} & $\xmark$ & $\xmark$ & $\cmark$ & $\min_{\theta} \E_{p(\mathcal{D}^T)}[\ell (\mathcal{D}^T; \theta)]$\tnote{2} \\
Multi-task Learning \cite{sener2018multi} & $\cmark$ & $\cmark$ & $\cmark$ & $\min_{\theta^{sh}, \theta^{\tau}} \E_{p(\mathcal{T})} \E_{p(\mathcal{D}_{\tau})} [\ell(\mathcal{D}_{\tau}; \theta^{sh}, \theta^{\tau})]$\tnote{3}\\
Meta-Learning \cite{wang2024simple}  & $\cmark$ & $\xmark$ & $\cmark$ & $\min_{\theta} \E_{p(\mathcal{T})}\E_{p(\mathcal{D}_{\tau})}[\ell (\mathcal{D}_{\tau}^{tr}, \mathcal{D}_{\tau}^{te}; \theta)], \;  \mathcal{D}_{\tau} = \mathcal{D}_{\tau}^{tr} \cup \mathcal{D}_{\tau}^{te}$\tnote{4} \\
Domain Adaptation \cite{farahani2021brief} & $\xmark$ & $\cmark$ & $\cmark$ & $\min_{\theta} \E_{p(\mathcal{D}^T)}[\ell (\mathcal{D}^T; \theta)]$ \\
Homogeneous DG \cite{zhou2022domain} & $\xmark$ & $\cmark$ & $\xmark$ & $\min_{\theta} \E_{p(\mathcal{S})}\E_{p(\mathcal{D}_s)} [\ell(\mathcal{D}_s; \theta)]$\tnote{5} \\
Heterogeneous DG \cite{wang2020heterogeneous, shu2021open} & $\xmark$ & $\xmark$ & $\xmark$ &  $\min_{\theta} \E_{p(\mathcal{S})}\E_{p(\mathcal{D}_s)} [\ell(\mathcal{D}_s; \theta)]$ \\
\hline
\end{tabular}
\begin{tablenotes}
    \item[1] In this table, we use $\theta$ to denote the set of all trainable parameters in the model. 
    \item[2] $\mathcal{D}^T$ represents the target domain.
    \item[3] $\theta^{sh}$ and $\theta^\tau$ denote the shared and the task-specific parameters, respectively.
    \item[4] $\mathcal{D}_{\tau}^{tr}$ and $\mathcal{D}_{\tau}^{te}$ indicate the meta-train and meta-test splits of task $\tau$'s data, respectively.
    \item[5] $\mathcal{D}_s$ and $\mathcal{S}$ represent a dataset from source domain $s$ and a set of source domains, respectively. 
\end{tablenotes}
\end{threeparttable}
}
\end{table*}

\subsection{Meta-learning for Domain Generalization: Promises and Challenges}
Meta-learning is considered a promising approach for DG, as it trains models to build transferable knowledge that can be efficiently applied to different tasks and domains. This aligns with the goal of DG to generalize to unseen domains. By training on a variety of learning tasks, meta-learning enables models to acquire knowledge that is transferable across different domains. In other words, meta-learning enables models to learn how to learn, which significantly enhances their ability to adapt to new and unseen domains by leveraging previously acquired knowledge. This adaptability is crucial for DG, where the goal is to generalize beyond the training domains (see Table~\ref{table:meta} for a comparison between traditional machine learning and meta-learning).
In meta-learning, models are trained episodically where each task or episode simulates a domain shift. This procedure effectively equips the model with the ability to handle domain shifts by encouraging the development of features that are robust to the variations between training and test domains. By repeatedly encountering a variety of tasks, the model learns to extract features that are crucial for performance across different domains, rather than overfitting to the idiosyncrasies of any single domain. These robust features are more likely to be relevant in unseen domains, thus improving the overall generalization of the model.
Additionally, the intrinsic sample efficiency of meta-learning can lead to more effective utilization of limited data. Meta-learning is designed to achieve high performance with fewer examples by leveraging the knowledge gained from previous tasks. This sample efficiency is particularly beneficial in scenarios where data from new domains is scarce or expensive to obtain.

One of the main challenges of using meta-learning for DG is ensuring a sufficient diversity of learning tasks during the meta-training phase. Without enough diversity, the model may lack the capacity to generalize effectively to completely new domains.
Additionally, the degree of the distributional shift between the training domains and the test domain can be substantial. If the unseen test domain differs significantly from the training domains, the model may encounter difficulties in generalizing.
Furthermore, while meta-learning is designed to be sample-efficient, in reality, it often requires a large number of tasks to achieve high performance, which might not always be feasible or accessible.
It is also important to note that effective domain generalization often requires understanding causal relationships within the data, which meta-learning algorithms might not capture without proper inductive biases or domain knowledge integration.

\begin{table*}[!ht]
\footnotesize
\setlength\extrarowheight{1pt} 
\caption{This table outlines the key differences between traditional machine learning and meta-learning across various aspects such as tasks, training datasets, purpose, loss functions, weights, and the role of these weights. The comparison highlights how meta-learning enhances adaptability and generalization to new and unseen domains by leveraging knowledge from a variety of tasks, in contrast to traditional machine learning which typically focuses on optimizing performance for a specific task.}
\label{table:meta}
\begin{tabularx}{\textwidth}{C|C|C}
\hline
 & \textbf{Machine Learning} & \textbf{Meta-Learning}  \\ \hline \hline
\textbf{Task} & \multicolumn{2}{c}{Image recognition, semantic analysis of text} \\
\hline
\textbf{Training Dataset} &	Typically a large dataset annotated for the current task & Variety of tasks, often organized into training episodes. Each episode contains a small amount of data from a particular task.\\
\hline
\textbf{Purpose} & 	Learn to perform a specific task from a dataset & 	Learn to perform specific tasks and also adapt to new tasks using only a few examples or new datasets (domains) with unseen distributions. \\
\hline
\textbf{Loss Function} & A mathematical function that measures the discrepancy between the predicted output and the ground truth output & Measures not only the prediction error on the current task (same as machine learning loss) but also optimizes adaptability to new tasks.\\
\hline
\textbf{Weights} & Learned parameters that are adjusted during the training process to minimize the loss on that task & Base-level weights which are task-specific, and meta-level weights or hyperparameters that guide the learning process across tasks and datasets (different domains).\\
\hline
\textbf{What are these weights good for} & Transforming the raw features into a more representative space (for example, a more linearly separable space for classification tasks) & Should be able to ignore features only relevant to domain differences (robust against domain shift), focusing on the features relevant to the task. \\
\hline
\end{tabularx}
\end{table*}

\subsection{Formalization of Meta-Learning for Domain Generalization}
\label{subsec:formalization}
In meta-learning, the goal is to solve new unseen tasks by leveraging the knowledge gained from solving $N$ previous tasks. Each task $\mathcal{T}_i$ corresponds to a dataset $\mathcal{D}_i = \{(x^i_{j}, y^i_{j})  \, |   \, x_j \in \mathcal{X},\, y_j \in \mathcal{Y}\}_{j=1}^{n}$, where $\mathcal{D}_i$ is divided into a support set $\mathcal{D}_i^\mathcal{S}$ and a query set $\mathcal{D}_i^\mathcal{Q}$. It is important to note that in meta-learning, all tasks are sampled from a unique task distribution $p(\mathcal{T})$. 

In domain generalization, we are given data from $M$ 
different source domains and aim to perform well on a new unseen domain, known as the target domain. Specifically, each domain $d_i$ is defined as follows:
 \begin{equation} 
    \label{eq:domain}
    \begin{split}
        d_i \triangleq \{p_i(x), p_i(y|x)\},
    \end{split}
\end{equation}
where $p_i(x)$ refers to the distribution of the input data for domain $d_i$, and this distribution may vary across different domains. The term $p_i(y|x)$ represents the conditional probability distribution of the target $y$ given the input data $x$. In a homogeneous DG setting, $p_i(y|x)$ is assumed to be consistent across all domains, meaning the label space is the same. However, in a heterogeneous DG setting, $p_i(y|x)$ can vary, as there can be disjoint label spaces across different domains. 

For instance, consider the PACS~\cite{li2017deeper} dataset for homogeneous DG; this dataset comprises images from 4 distinct domains: photo, art painting, cartoon, and sketch. Each domain shares the same set of 7 classes for labeling. Tasks are drawn from these varying domains, each with its unique statistical characteristics. Consequently, there is a significant domain shift between tasks, necessitating a model capable of generalizing across these different domains to effectively handle tasks from varied distributions. In contrast, the Visual Decathlon (VD)~\cite{rebuffi2017learning}  dataset includes 10 different domains, each encompassing a variety of categories, and is thus suitable for designing tasks for heterogeneous DG. While the aforementioned datasets focus on multi-source DG,  
where the model is trained across several source domains to provide representations that generalize to novel unseen domains, there are scenarios where only a single source domain is available. In these cases, the model is expected to generalize to target domains without prior exposure to multiple source domains. This challenge is known as single-domain generalization. Additionally, some problems require the model to handle an open-set of target domains as opposed to a limited or closed-set of target domains. This requirement significantly increases the difficulty of domain generalization problems.

As it can be conceived, a domain can be viewed as a special case of a task from the perspective of meta-learning. Therefore, the meta-learning paradigm can also be applied to domain generalization problems, where each domain is treated as if it were a distinct task, defined by a specific source $\mathcal{S}_i$ and associated with a dataset $\mathcal{D}_i$. It's also worth mentioning that we can leverage zero-shot learning capabilities for new unseen domains when integrating domain generalization techniques with meta-learning.

\vspace{7pt}

\section{Taxonomy}
\label{sec:taxonomy}
To advance DG through meta-learning, it is important to get an overview of the existing approaches and their similarities and differences. Here, we introduce a taxonomy and its key dimensions (axes) with which different meta-learning approaches can be analyzed. 
The taxonomy provides two axes, the discriminability axis, and the generalizability axis, representing how a model learns to generalize to unseen domains. This taxonomy aims to highlight the interaction and knowledge transferability between data points from different domains and classes. The end goal is to meta-learn domain-agnostic features and discriminative classifiers that perform well on unseen domains.

\begin{figure*}[ht!]
\centering
\includegraphics[width=1\linewidth]{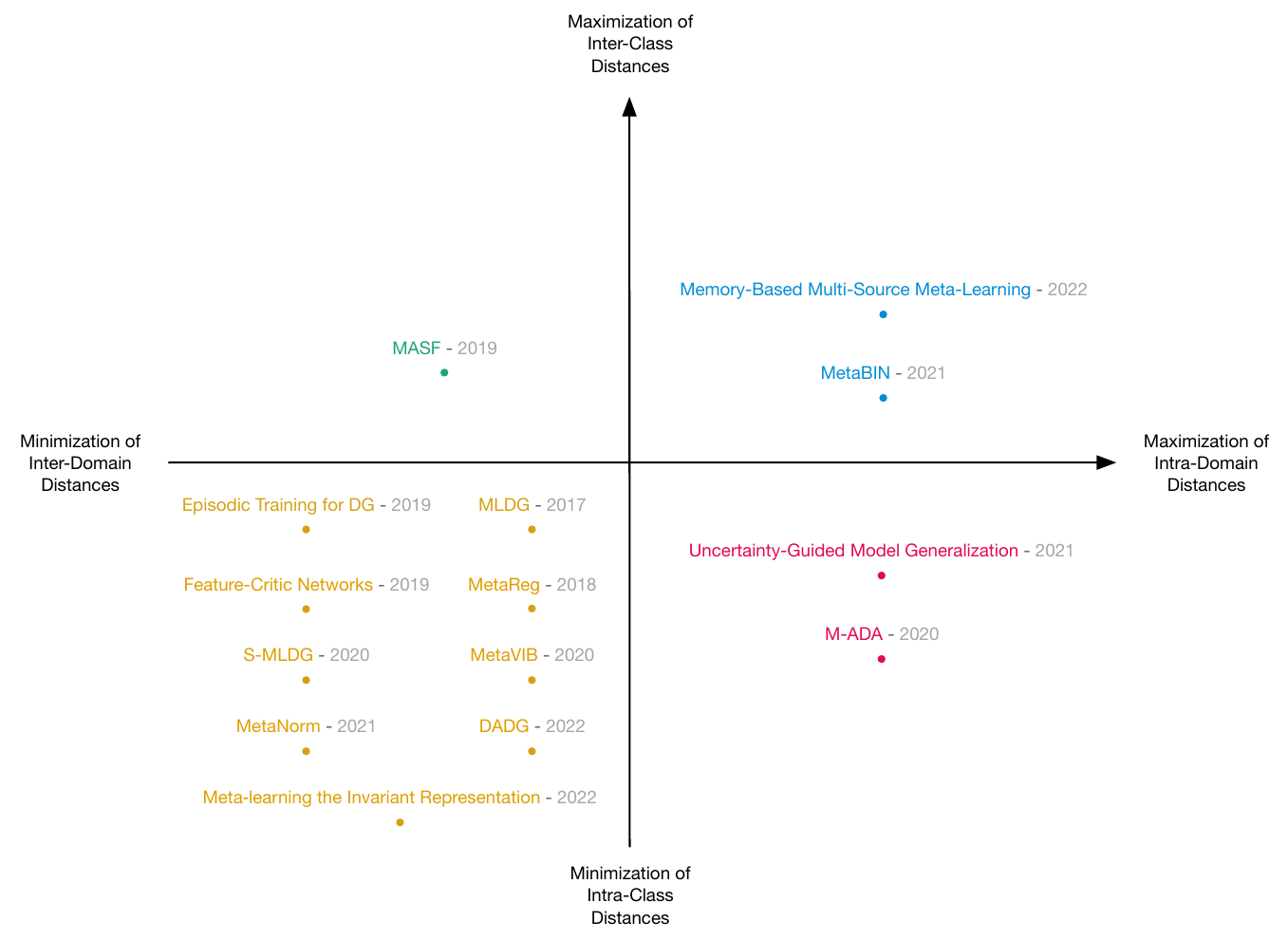}
\caption{An illustrative diagram of the meta-learning taxonomy for domain generalization. The quadrant chart highlights two principal axes: the first axis (the generalizability axis) represents the strategy of the feature extractor, contrasting the Minimization of Inter-Domain Distances with the Maximization of Intra-Domain Distances; the second axis (the discriminability aixs) depicts the classifier training process, distinguishing between the Minimization of Intra-Class Distances and Maximization of Inter-Class Distances. This diagram visually organizes domain generalization approaches, illustrating their distinct mechanisms for promoting generalization across unseen domains.}
\label{fig:taxonomy}
\end{figure*}

Compared to other meta-learning techniques, a unique characteristic of DG is its design principle that prevents the model from adapting to specific test domains. This is because, by definition, DG focuses on generalizing to completely unseen test domains, unlike domain adaptation, where the model has some form of access or exposure to information from the test domain during training. 
This distinct feature requires the meta-learning approach to \emph{generalize} instead of \emph{adapt}. Consequently, it is crucial to balance the trade-off between domain-invariant and domain-specific features~\cite{chattopadhyay2020learning}. Specifically, in classification tasks, the domain-specific features that are learned must be carefully curated to enhance class separability even for unseen domains. 
The taxonomy presented in this survey is designed to address these aspects. We divide the categorization into two principal axes: the first concerning the treatment of domain features by the feature extractor (generalizability) and the second relating to the training methodology of the classifier within the model (discriminability).

\subsection{Generalizability Axis}
The feature extractor is a pivotal component in domain generalization, as it determines how the model perceives and processes data from diverse domains to learn generalized representations applicable across various settings. Approaches along this axis can be divided into two categories: (1) \textit{Minimization of Inter-Domain Distances} and (2) \textit{Maximization of Intra-Domain Distances}.

\subsubsection{Minimization of Inter-Domain Distances}
Some approaches aim to extract domain-invariant features by minimizing the distance between feature representations across different domains. The primary objective is to identify and harness the characteristics that are consistent across domains. In other words, these methods mainly focus on directly inducing invariant representations without diversifying inputs, concentrating on the consistent and discriminative aspects of the data. By doing so, they reduce the model's sensitivity to the idiosyncrasies of any given domain, thereby enhancing the robustness of the model against the shifts that occur in novel environments.

\subsubsection{Maximization of Intra-Domain Distances}
Alternatively, certain approaches seek to construct a feature extractor capable of handling a more extensive variety of data by maximizing inter-domain distances. These methods utilize strategies like data augmentation, noise injection, and domain randomization to expand the training data spectrum and encourage the model to learn features that are broadly applicable rather than domain-specific. By exposing the model to a more diverse set of modified inputs, these methods strive to replicate the variability that the model will encounter in real-world scenarios, thus promoting the learning of features with better generalizability. Through the explicit diversification of inputs from a domain, these approaches develop feature extractors prepared to handle greater variance in the data.

\subsection{Discriminability Axis}
The performance of DG models is highly dependent on the effectiveness of the classifier's training methodology. The strategies in this context are categorized into two groups based on how the classifier manages distances: (1) \textit{Minimization of Intra-Class Distances} and (2) \textit{Maximization of Inter-Class Distances}.

\subsubsection{Minimization of Intra-Class Distances}
Most approaches simply rely on minimizing the distances between instances within the same class. The intention is to encourage the model to group similar examples together. This method ensures that the model's predictions are consistent for similar instances.

\subsubsection{Maximization of Inter-Class Distances}
Some methodologies also explicitly maximize the distance between different classes by incorporating triplet loss to train the classifier. This not only minimizes inter-class distances but also explicitly maximizes intra-class distances. By doing so, the model is encouraged to learn robust representations that distinctly separate different classes. The use of triplet loss helps to develop a feature space where classes are well-delineated, resulting in clearly separated clusters. This enhances the classifier's discriminative power and its ability to generalize across domains.

Taken together, these two axes allow us to categorize domain generalization techniques based on how they train the feature extractor and classifier components to improve generalization. Methods can differ in whether they align or diversify representations across domains, as well as how they distinguish between classes during training, as illustrated in Figure~\ref{fig:taxonomy}.

\begin{figure}[h]
\centering
\includegraphics[width=1\linewidth]{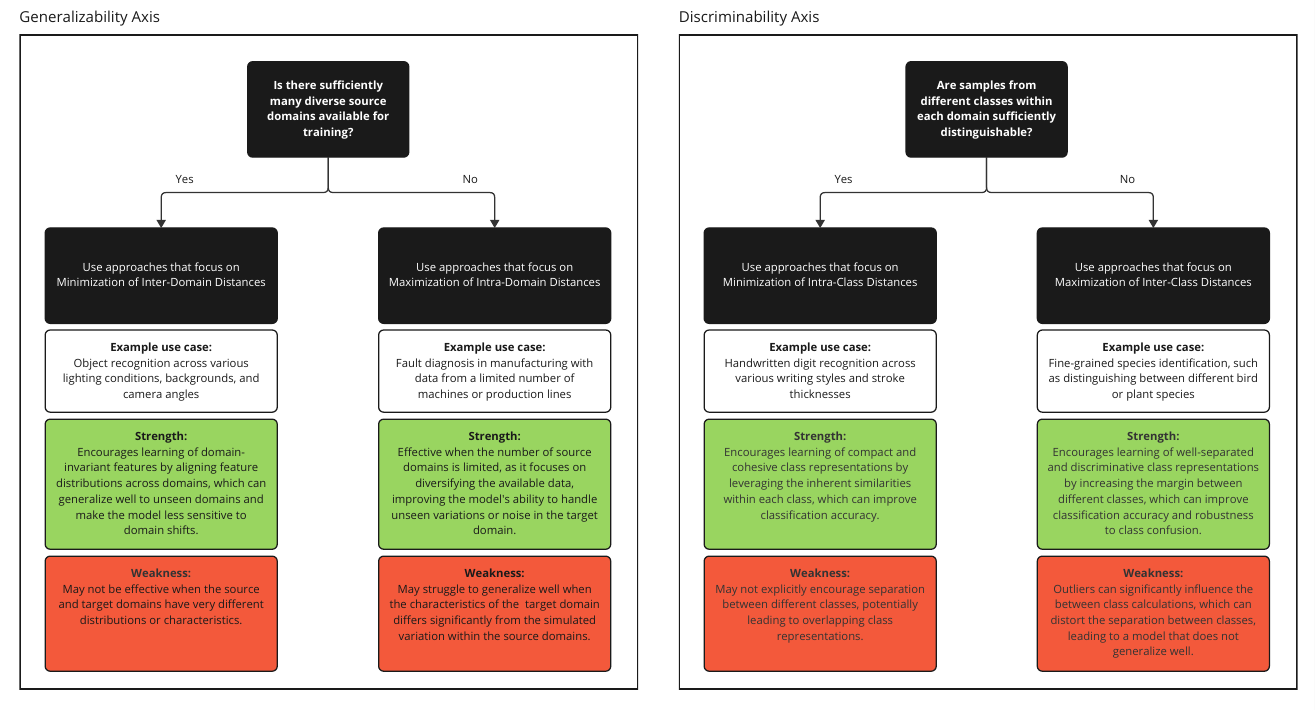}
\caption{A decision graph illustrating how to apply the taxonomy of meta-learning approaches for domain generalization. The decision graph categorizes techniques based on the two key aspects: the generalizability axis, which focuses on the feature extractor's strategy (minimizing inter-domain distances or maximizing intra-domain distances), and the discriminability axis, which focuses on the classifier's training process (minimizing intra-class distances or maximizing inter-class distances). This decision graph helps the reader navigate through the taxonomy, providing use cases, strengths, and weaknesses for each category to make the concepts more tangible, applicable, and actionable.}
\label{fig:decision}
\end{figure}

\paragraph{Application of the taxonomy} 
In Figure~\ref{fig:decision}, we present a decision graph to guide readers through the taxonomy to select the most appropriate meta-learning method for DG based on data availability and domain shifts. It categorizes meta-learning approaches based on the strategies employed for the feature extractor (generalizability axis) and the classifier training process (discriminability axis), facilitating an understanding of the relationships between various methods and their underlying principles.

\vspace{7pt}
\section{Methodologies}
\label{sec:methodologies}
In recent years, there has been significant interest in utilizing the meta-learning procedure to create models for DG. In this section, we will explore individual meta-learning methods for DG by describing the methodologies and motivations behind their designs. We discuss the strengths and weaknesses of each method in detail.

Meta-learning focuses on building a general model that can be used to adapt rapidly to new situations. For the first time, Li et al. exploited the idea behind meta-learning for DG by dividing the data from multiple source domains into non-overlapping meta-train and meta-test sets to simulate domain shifts in the training procedure. Therefore, we will discuss DG through meta-learning, starting with meta-learning for domain generalization (MLDG) and proceeding accordingly.

\subsection{MLDG}
Recent meta-learning studies have focused on learning good weight initializations for few-shot learning, particularly Model-Agnostic Meta-Learning (MAML)~\cite{finn2017model}. Li et al.~\cite{li2018learning} proposed MLDG (meta-learning for domain generalization), which draws inspiration from the MAML approach and trains models capable of performing well on OOD data by transferring knowledge across domains instead of tasks. It is important to note that DG assumes zero training examples from the target domain, in contrast to conventional meta-learning methods like MAML, which address few-shot scenarios using few training examples of new tasks. Through the meta-learning framework, MLDG will be exposed to domain shifts during training, enhancing its ability to handle domain shifts in various situations by learning general representations. Accordingly, they divide data from multi-source domains into meta-train and meta-test sets to simulate domain shifts. In the meta-train phase, parameters $\theta$ of the model are updated by gradient descent using the meta-train sets $\mathcal{S}_{tr}$. As a result, the following steps will be taken:
\begin{equation}
    \begin{split}
    \label{eq:rstdp}
        \nabla_{\theta} &= \ell^{\prime}_{\theta}(\mathcal{S}_{tr}; \theta),\\ 
        \theta^{\prime} &= \theta - \alpha \nabla_{\theta},
    \end{split}
\end{equation} 
where $\alpha$ is the meta-train (inner loop) learning rate. Following that, in the meta-test phase, we determine the loss using the meta-test sets (virtual-test domains) $\mathcal{S}_{te}$ with the parameters obtained in the meta-train phase. Consequently, this phase involves calculating gradients over gradients to update the model's parameters. The parameters update can be formulated as: 
\begin{equation}
    \begin{split}
    \label{eq:rstdp}
        \theta = \theta - \beta \frac{\partial (\ell(\mathcal{S}_{tr}; \theta) + \gamma \ell(\mathcal{S}_{te}; \theta^{\prime}))}{\partial \theta},
    \end{split}
\end{equation} 
where $\ell$ is the loss function ,and $\beta$ is the meta-test (outer loop) learning rate, and $\gamma$ weights meta-train and meta-test.

MLDG is built on top of MAML and improves the generalization of MAML, which is model-agnostic and easily applicable to reinforcement learning (RL) problems. However, Like MAML, it is computationally expensive since it requires the calculation of second-order derivatives.

\subsection{MetaReg}
Balaji et al.~\cite{balaji2018metareg} introduced MetaReg (domain generalization using meta-regularization) that incorporates a regularization term trained using meta-learning to encourage models to extract domain-invariant features and reduce sensitivity to domain-specific variations to train models that are robust to domain shifts.
Their proposed model consists of a feature network $\psi$ along with $p$ task networks $\theta_i$ for $p$ source domains. The feature network is designed to extract more general features, and the tasks networks are utilized to enforce domain-specificity in the networks so that they can apply regularizers to make them domain-invariant. Therefore, they implement the meta-learning framework to train the regularizer term and incorporate it into the loss function. The regularizer term will be trained using every pair of source domains $(a, b)$. This procedure can be expressed by the following set of equations:
\begin{equation}
    \begin{split}
    \label{eq:metareg-inner}
        &\beta^{1} \leftarrow \theta_{a}^{(k)}\\
        &\beta^{t} = \beta^{t-1} - \alpha \nabla_{\beta^{t-1}}[L^{(a)}(\psi^{(k)}, \beta^{t-1}) + R_{\phi}(\beta^{t-1})] \quad  \\
        & ~~~~\forall t \in \{2,\ldots,l\}\\
        & \hat{\theta}_{a}^{(k)} = \beta^{l}
    \end{split}
\end{equation} 

\begin{equation}
    \begin{split}
    \label{eq:metareg-outer}
        &\phi^{(k+1)} = \phi^{(k)} - \alpha \nabla_{\phi} L^{(b)}(\psi^{(k)},\hat{\theta}_{a}^{(k)})|_{\phi = \phi^{(k)}}
    \end{split}
\end{equation} 

Where Eq. \eqref{eq:metareg-inner} describes the inner loop of the meta-learning paradigm that performs $l$ steps of gradient descent using meta-train set on a new task network with parameters $\theta_{a}^{(k)}$ which is the base model's task network parameters of the $a^{th}$ domain at iteration $k$ to obtain $\hat{\theta}_{a}^{(k)}$. In addition, Eq. \eqref{eq:metareg-outer} refers to the outer loop of the meta-learning strategy, which incorporates the meta-test set to minimize the unregularized loss with parameters $\hat{\theta}_{a}^{(k)}$ with respect to the regularizer parameter $\phi$. By using the trained regularizer, we fine-tune a new model on the entire source dataset. To further add, weighted $L_1$ loss is used as the regularization function, i.e., $R_{\phi}(\theta) = \sum_i \phi_i \lvert\theta_i\rvert$ to build a learnable weight decay mechanism and diminish domain-specific characteristics in task networks.

Although MetraReg increases the training complexity and requires several domains for practical training, it is easily applicable to deep neural networks and helps reduce overfitting.

\subsection{Feature-Critic Networks}
\begin{figure}[ht]
\centering
\includegraphics[width=0.9\linewidth]{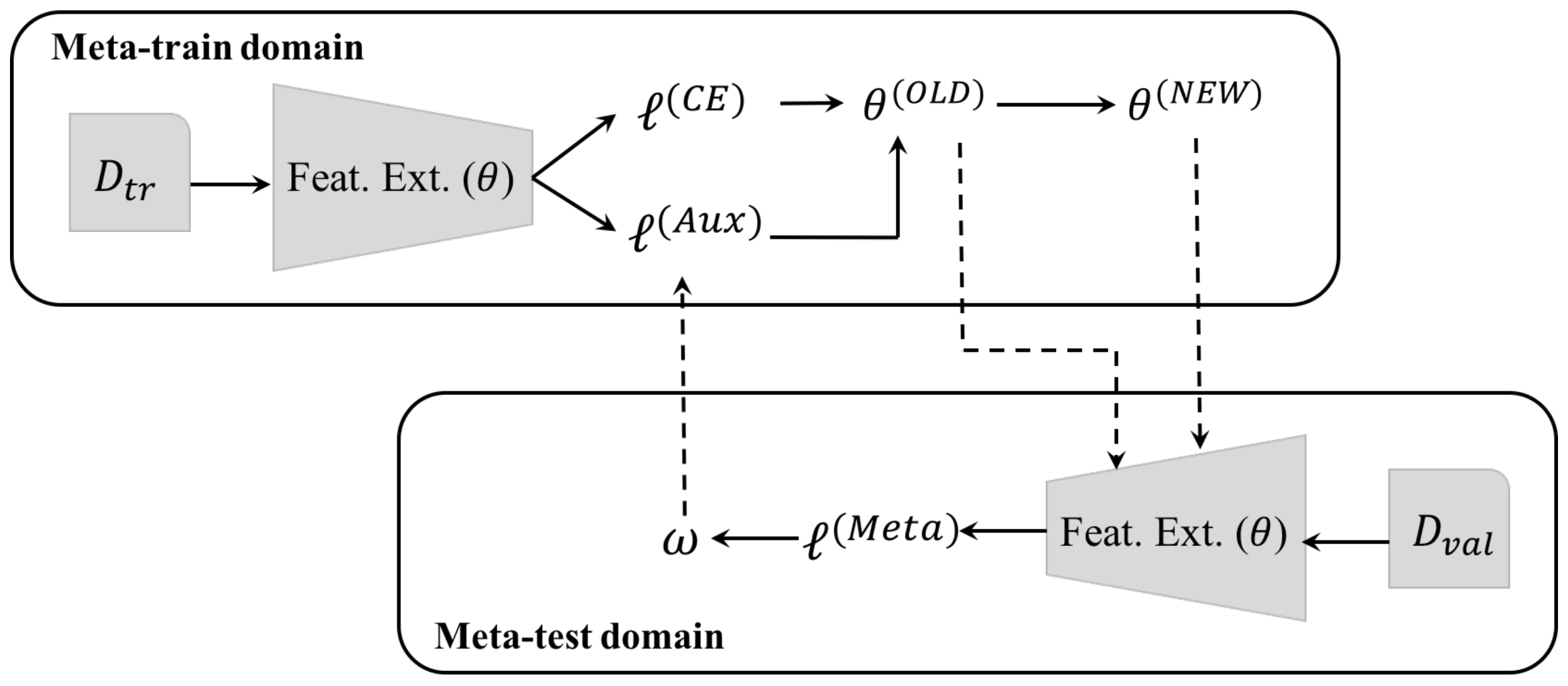}
\caption{Visual representation of the Feature-Critic Networks for heterogeneous domain generalization, depicting the base network's feature extraction guided by an auxiliary loss from the feature-critic networks to promote domain-invariant feature extraction. The domain-invariant features are generated by minimizing the inter-domain distances.}
\label{fig:feature-critic}
\end{figure}

Feature-Critic Networks introduced by Li et al.~\cite{li2019feature} is tailored for heterogeneous domain generalization by drawing inspiration from the meta-learning framework. The proposed model meta-learns a loss function that promotes domain robustness. In this way, they train a robust feature extractor that can be used with any classifier for addressing various problems from different domains. More specifically, they introduced a learnable auxiliary loss $\ell_{\omega}\textsuperscript{(Aux)}$ resulting from optimizing the feature-critic network $h_{\omega}$ to encourage the base network to extract domain agnostic features, as illustrated in Figure~\ref{fig:feature-critic}.

Based on the following equations, parameters $\theta$ of the base network is computed using the meta-train data $\mathcal{D}_{tr}$:
\begin{equation}
    \begin{split}
    \label{eq:feature-critic-inner}
        \theta\textsuperscript{(OLD)} &= \theta - \alpha \nabla_{\theta} \ell\textsuperscript{(CE)}(\mathcal{D}_{tr} | \theta),\\
        \theta\textsuperscript{(NEW)} &= \theta - \alpha \nabla_{\theta} (\ell\textsuperscript{(CE)}(\mathcal{D}_{tr} | \theta) + \ell_{\omega}\textsuperscript{(Aux)}(\mathcal{D}_{tr} | \theta)),
    \end{split}
\end{equation} 
where $\ell\textsuperscript{(CE)}$ is the cross-entropy loss, and the auxiliary loss $\ell_{\omega}\textsuperscript{(Aux)}$ is defined as follows: 
\begin{equation}
    \begin{split}
    \label{eq:aux-loss}
        \ell_{\omega}\textsuperscript{(Aux)} := \frac{1}{N} \sum_{i=1}^{N} \ softplus(h_{\omega}(f_{\theta}(x_i))),
    \end{split}
\end{equation} 
where $f_{\theta}$ is the feature extractor. Also, the softplus function is applied to ensure the output is non-negative. 

In order to determine parameters $\omega$ of the feature-critic network, the following optimization will be used with the meta-test data $\mathcal{D}_{val}$:
\begin{equation}
    \begin{split}
    \label{eq:feature-critic-outer}
        \min_{\omega} \ tanh(\ell\textsuperscript{(CE)}(\mathcal{D}_{val} | \theta\textsuperscript{(NEW)}) - \ell_{\omega}\textsuperscript{(CE)}(\mathcal{D}_{val} | \theta\textsuperscript{(OLD)})).
    \end{split}
\end{equation} 

The tanh function can be considered a softer version of gradient clipping in this context.

Ultimately, we can utilize $g_{\phi} \circ f_{\theta}$, where the trained base network $f_{\theta}$ will serve as a fixed feature extractor for target domains, and $g_{\phi}$ would be the classifier. For heterogeneous DG, we have $N$ distinct classifiers, denoted as $g_{\phi_1}, g_{\phi_2}, \ldots, g_{\phi_N}$. On the other hand, in the homogeneous DG, we have a single classifier $g_{\phi}$ that can be shared across all domains.

Feature-critic networks can effectively deal with heterogeneous DG where label spaces differ. Nevertheless, there is no guarantee that the gradients of the feature-critic networks will not conflict with those from the supervised loss, which could negatively affect the performance of the model.

\subsection{Episodic Training for DG}
\begin{figure}[ht]
\centering
\includegraphics[width=0.8\linewidth]{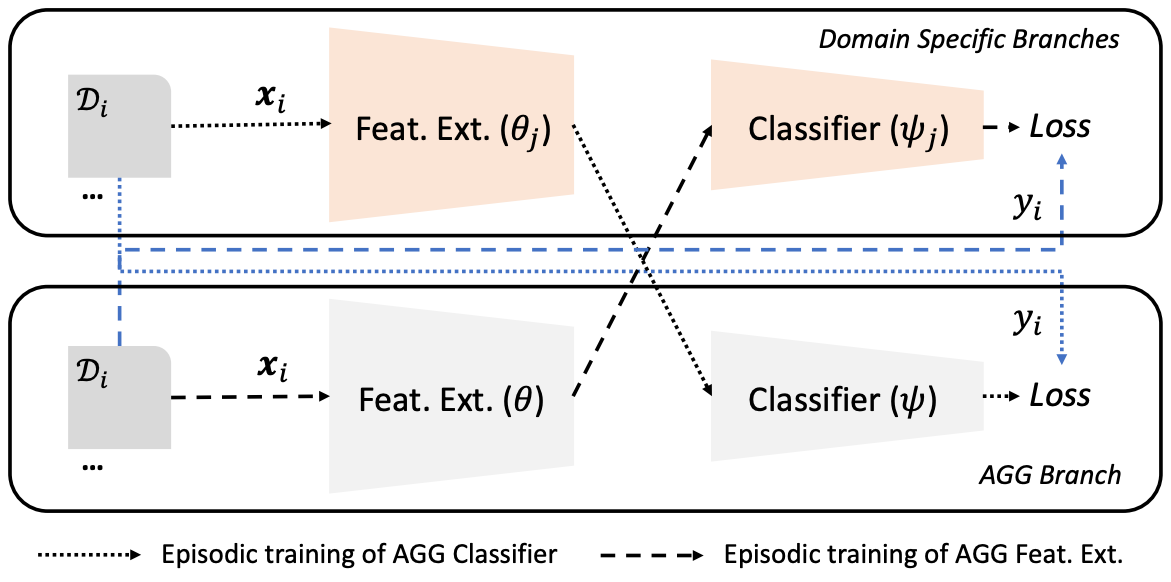}
\caption{Overview of Episodic Training for DG framework, illustrating the regularization process where a feature extractor is trained with classifiers from different domains and vice versa, promoting out-of-distribution robustness.}
\label{fig:episodic}
\end{figure}

A novel framework for DG is proposed by Li et al.~\cite{li2019episodic} based on episodic training that resembles the meta-learning training approach. In particular, they used episodic training to train a deep neural network, decomposed into feature extractor and classifier components. Each component is trained by simulating interactions with a poorly tuned partner for the current domain. Moreover, to construct episodes, the framework uses data from domain $A$ to be processed by the classifier trained on domain $B$, which has not been exposed to data from $A$, and vice versa. This cross-domain exposure ensures that each component becomes sufficiently robust at handling OOD data, as depicted in Figure~\ref{fig:episodic}. 

The intuition of the episodic training using a mismatched classifier is to regularize the feature extractor by asking it to produce features robust enough that even an arbitrary classifier will be able to perform properly. Accordingly, three episodic training strategies are introduced: 1) Train feature extractor to support classifiers from other domains; 2) Train classifier to accept features from other domains; 3) Train feature extractor to support a randomly initialized classifier.

The proposed approach requires only a simple modification of the training procedure and does not rely on special data augmentation or optimization algorithms. It also allows for heterogeneous label spaces across domains since no classifier needs to be shared. However, there is a risk that the feature extractor could learn degenerate representations if the randomness imposed overwhelms the true objectives of the training.

\subsection{Meta‑learning the Invariant Representation}
\begin{figure}[ht]
\centering
\includegraphics[width=1\linewidth]{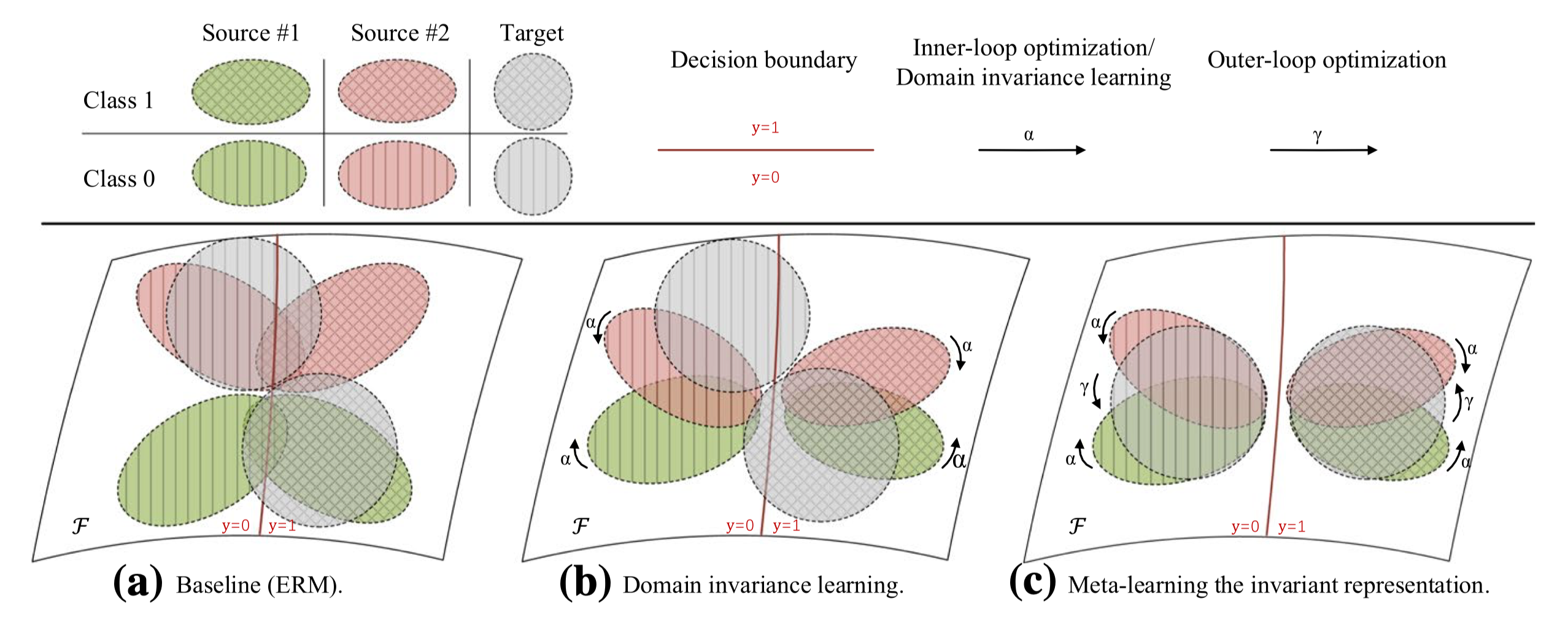}
\caption{Depiction of the meta-learning for invariant representation method. \textbf{(a)} Compared to the ERM baseline, \textbf{(b)} domain invariance learning decreases the discrepancy among source domains and excels in source-domain classification, yet it may still result in significant errors on the target domain. \textbf{(c)} The proposed approach employs bilevel meta-learning to further minimize the discrepancy between the target and source domains, enabling a hypothesis learned from the source domains to generalize effectively to the target domain.}
\label{fig:invariant}
\end{figure}

Learning invariant representations across source domains has shown effectiveness for DG. However, overfitting to source domains can limit generalization to a target domain that differs greatly from the sources. Jia et al.~\cite{jia2022meta} proposed a meta-learning algorithm via bilevel optimization to improve the out-of-domain robustness of the learned invariant representation. 
Most meta-learning algorithms share a common limitation: they use task objectives directly as the inner-loop and outer-loop objectives. This approach can be suboptimal because it is highly abstracted from the feature representation. To address this issue, the authors focus on a meta-learning approach aimed at reducing the discrepancy between the target domain and source domains, where in each training iteration, any domain can become the meta-target domain while the rest serve as the meta-source domains. Specifically, they developed a bilevel meta-learning procedure based on the first-order MAML framework, which achieves high computational efficiency.
The paper employs a $\mathcal{Y}$-discrepancy measure~\cite{zhang2012generalization} to quantify domain discrepancy, effectively capturing both covariate and conditional shifts between domains. Notably, $\mathcal{Y}$-discrepancy has been utilized in previous research for domain invariance learning~\cite{zhang2021quantifying}. The primary goal of this algorithm is to minimize the $\mathcal{Y}$-discrepancy between the source domains and the target domain.
A gradient-based meta-learning algorithm is provided to solve the bilevel optimization problem efficiently. The algorithm alternates between sampling meta-tasks, updating the feature embedding and classifiers on the meta-training set, and updating the meta-parameters based on the meta-test performance.
The inner-loop objective aims to minimize discrepancy across different source domains, while the outer-loop objective aims to minimize discrepancy between source domains and a potential target domain. In particular, adversarial learning is employed to estimate and minimize the $\mathcal{Y}$-discrepancy between domains in both the inner and outer loops of the bilevel meta-learning algorithm.
In essence, this meta-learning approach reduces the discrepancy between the target domain and the convex hull of the source domains as depicted in Figure~\ref{fig:invariant}. This enables learning a robust invariant representation for improved domain generalization.

The introduced model learns a robust invariant representation that generalizes well to unseen domains by optimizing the feature embedding to minimize the discrepancy between meta-source and meta-target domains during training. However, like other domain generalization methods, the effectiveness of the proposed approach may depend on the diversity of the available source domains. If the source domains lack diversity, the learned invariant representation may not generalize well to significantly different target domains.

\subsection{MASF}
\begin{figure}[ht]
\centering
\includegraphics[width=1\linewidth]{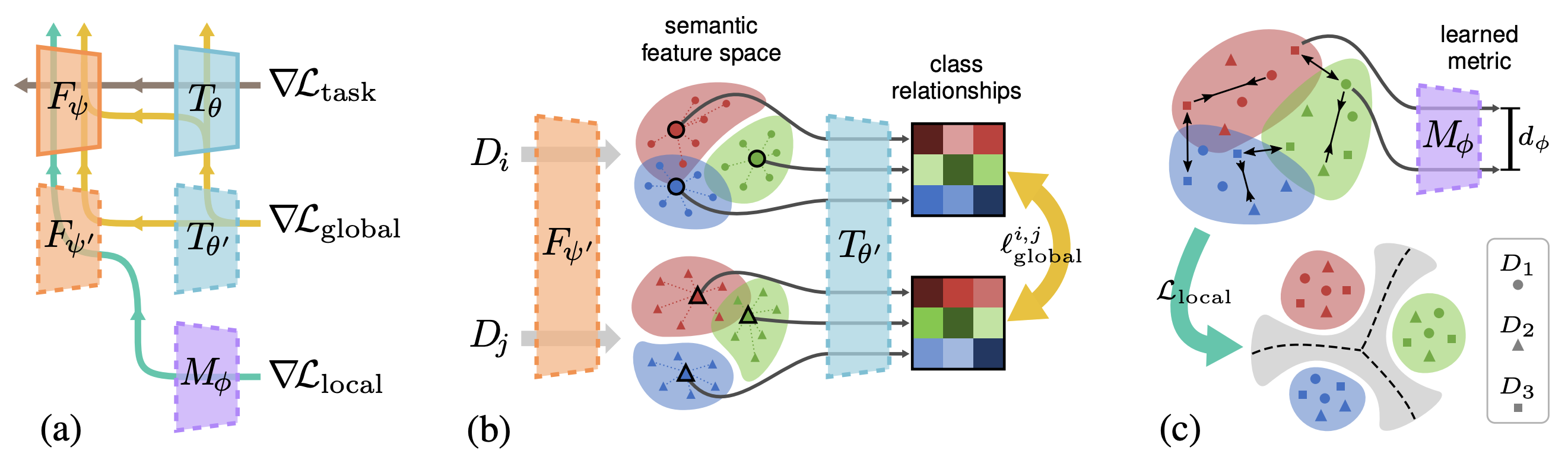}
\caption{Overview of the MASF approach for Domain Generalization, $F_{\psi}$, and $T_{\theta}$ represent the feature extractor and task network, respectively, with $F_{\psi^{\prime}}$ and $T_{\theta^{\prime}}$ being their updated versions after inner gradient updates on the task loss $\mathcal{L}_{task}$. $M_{\phi}$ is the metric embedding network, and $D_{k}$ represents different source domains. (a) Illustration of gradients flow of episodic training with domain shift simulation; (b) Global alignment for consistent class relationships across domains; (c) Local sample clustering to promote the Maximization of Inter-Class Distances.}
\label{fig:semantic}
\end{figure}

Model-agnostic learning of semantic features (MASF) has been presented by Dou et al.~\cite{dou2019domain}, in which two complementary losses that explicitly regularize the semantic structure of the feature space are introduced. This approach aims to enhance domain generalization within the meta-learning framework by splitting source domains into meta-train and meta-test sets to simulate domain shift through an episodic training procedure. In more detail, MASF suggests enforcing semantic features through global class alignment and local sample clustering, as illustrated in Figure~\ref{fig:semantic}. These supplementary losses are explicitly derived in an episodic learning procedure. 

The objective of global class alignment is to structure the feature space such that it preserves the learned inter-class relationships on unseen data via explicit regularization. By ensuring consistent class relationships, knowledge transfer across domains is facilitated, thus improving domain generalization. This inter-class alignment is done by exploiting what the model has learned about class ambiguities in the form of per-class soft labels and enforcing their consistency within each two distinct domains by minimizing the symmetric KL divergence between their soft confusion matrix created from the collection of soft labels.

The objective of local regularization is to enhance robustness by promoting feature compactness, grouping features of samples within the same class closely together while being distinct from features of different classes. This is crucial to prevent ambiguous decision boundaries and sensitivity to unseen domain shifts. As a result, an embedding network is utilized to process the extracted features. The output of this network is then used to measure the distance between the input features of randomly selected pairs of samples from all domains. The training procedure employs either contrastive loss or triplet loss.

The proposed model captures inter-class relationships and ensures intra-class compactness, improving the generalization of the model; however, applicability to large-scale problems with many classes may be a challenge.

\subsection{S-MLDG}
\begin{figure}[ht]
\centering
\includegraphics[width=1\linewidth]{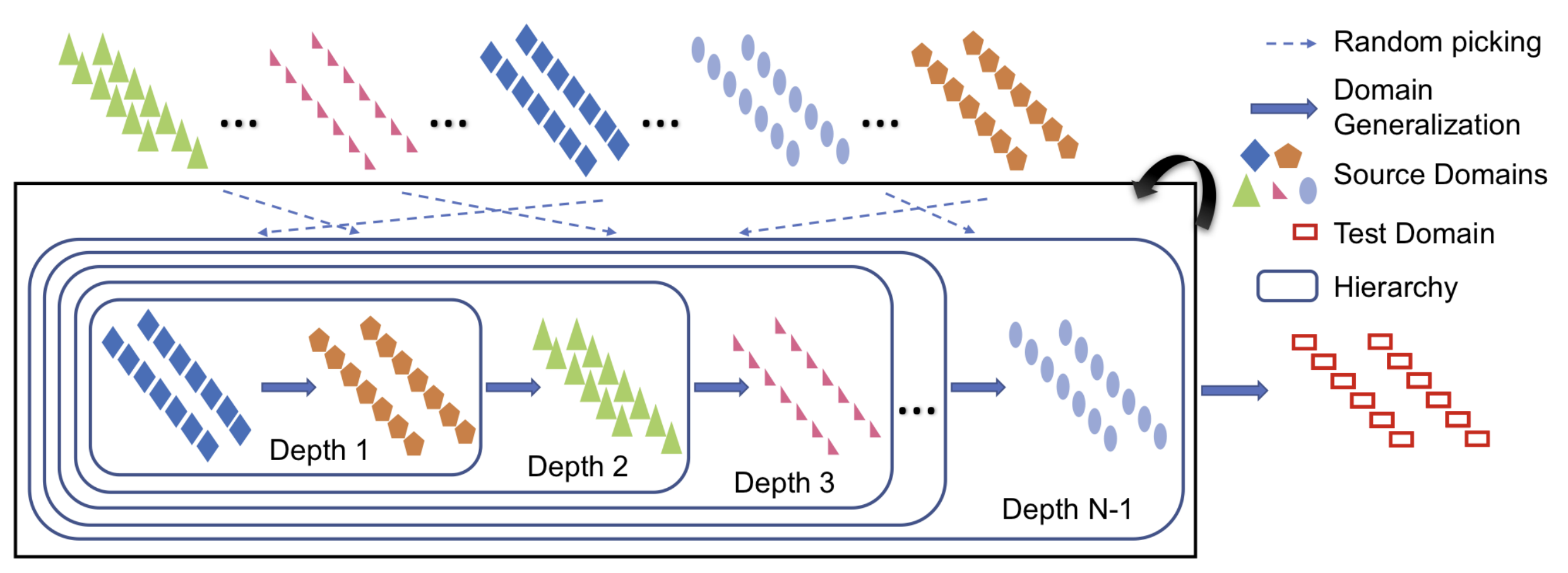}
\caption{Illustration of the S-MLDG training procedure, showing the sequential training across multiple domain trajectories with random sequences to simulate lifelong learning and increase the diversity of domain-shift episodes.}
\label{fig:smldg}
\end{figure}

Li et al.~\cite{li2020sequential} proposed a framework for sequential learning for domain generalization (S-MLDG) based on the MLDG to enhance its performance by incorporating sequential learning and lifelong learning. They train a base DG model sequentially along a trajectory spanning the source domains, optimizing the cumulative performance across the entire trajectory instead of individual steps. In the case of MLDG, it recursively applies MLDG along the trajectory, simulating lifelong DG learning, as shown in Figure~\ref{fig:smldg}. This approach encompasses more unique DG episodes compared to the base algorithms. While MLDG considers N distinct domain-shot episodes, this method considers N! distinct domain-shift episodes. Furthermore, the approach defines a loss function that aggregates the performance on each step of the trajectory through domains. Additionally, it incorporates shuffling domain orders and sampling new batches for each iteration, ensuring diversity and preventing overfitting.

The fast first-order approximation, denoted as FFO-S-MLDG, constitutes a streamlined variant of the S-MLDG algorithm, drawing inspiration from Reptile to mitigate its computational overhead. FFO-S-MLDG uses a first-order approximation of the original S-MLDG algorithm, enabling it to be trained more quickly and efficiently.

Although the training complexity is higher compared to MLDG due to the optimization over domain trajectories, this approach consistently demonstrates improvements over strong baselines on multiple benchmarks. This is due to its ability to generate more unique DG episodes.

\subsection{MetaVIB}
\begin{figure}[ht]
\centering
\includegraphics[width=1\linewidth]{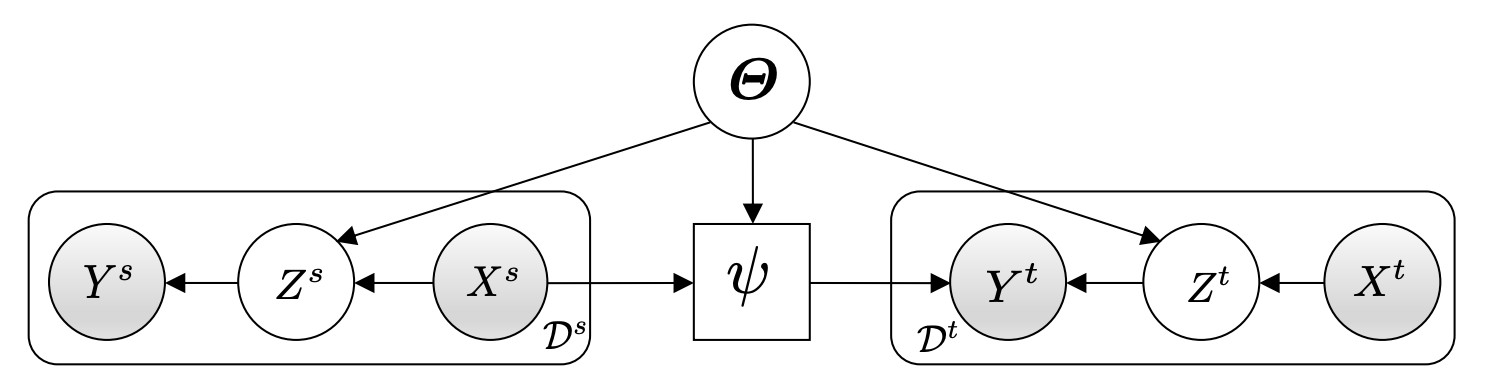}
\caption{Computational graph of the MetaVIB model for domain generalization, depicting the training process where domain-invariant representations are learned. Global parameters ($\Theta$) and classifier parameters ($\psi$) are optimized over source domains, which are split into meta-train ($\mathcal{D}^{s}$) and meta-test ($\mathcal{D}^{t}$) sets to simulate domain shifts within each episodic training phase. During the test phase, $\Theta$ generates representations for the target domain, while $\psi$ is used for predictions of data from the source domain. While this is a typical example of the inter-domain minimization and intra-class minimization paradigm, the distinctive feature of MetaVIB lies in its explicit probabilistic modeling approach. }
\label{fig:vib}
\end{figure}

Meta variational information bottleneck (MetaVIB) has been introduced by Du et al.~\cite{du2020learning} as a probabilistic meta-learning model to handle uncertainty and domain shifts for domain generalization by learning domain-invariant representations using variational auto-encoders. In order to address the uncertainty associated with predictions on the unseen target domain, the model represents classifier parameters as distributions inferred from source domains. In a similar manner to other algorithms, source domains are split into meta-train and meta-test sets in each training episode to simulate domain shift, as illustrated in Figure~\ref{fig:vib}. MetaVIB is derived from novel variational bounds of mutual information by leveraging the meta-learning setting. Also, the meta-learning objective maximizes performance on meta-test while minimizing domain divergence via MetaVIB. As a result, MetaVIB learns to gradually narrow domain gaps to achieve domain invariance, while maximizing prediction accuracy through episodic training. This encourages learning efficient representations that preserve information for prediction but remove domain-specific statistics, thereby enabling the learning of domain-agnostic representations.

MetaVIB explicitly addresses the uncertainty associated with predictions for new domains through probabilistic modeling. Also, by limiting the information flow from the inputs to only what is relevant for the classification task through the information bottleneck. This prevents the model from encoding irrelevant domain-specific details. However, it does not explicitly align distributions across domains, implying a certain level of domain shift might remain.

\subsection{M-ADA}
\begin{figure}[ht]
\centering
\includegraphics[width=0.9\linewidth]{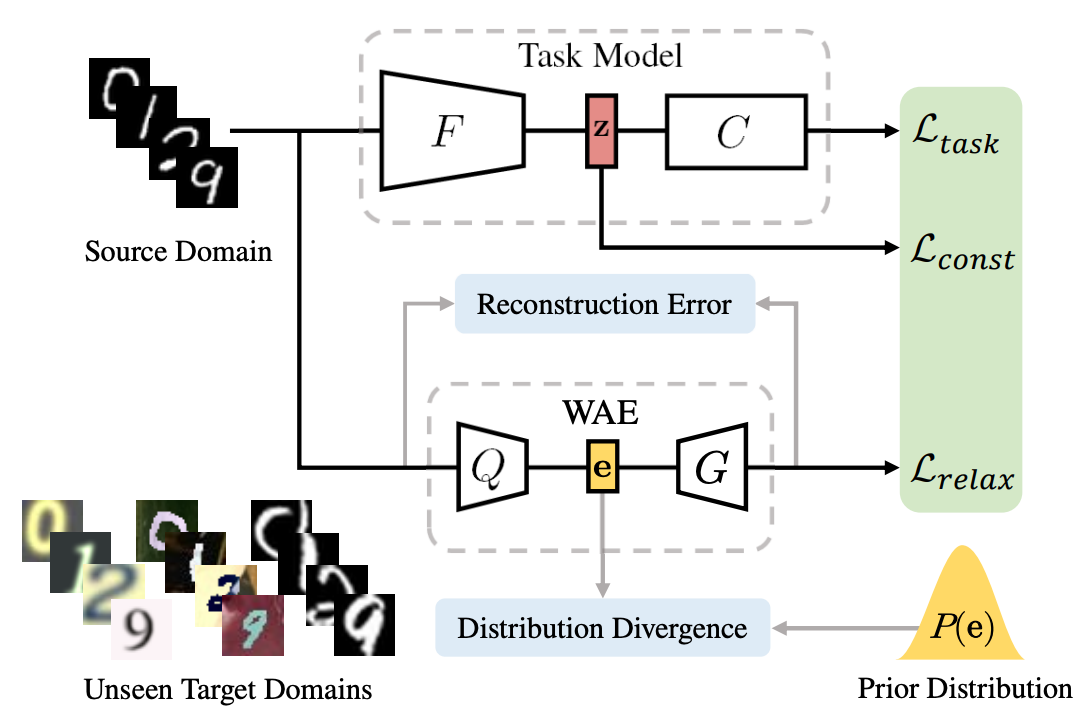}
\caption{Overview of M-ADA method for single domain generalization using adversarial domain augmentation. The architecture employs a task model comprising a feature extractor $F$ and classifier $C$, alongside a Wasserstein Autoencoder (WAE), to generate fictitious domains.}
\label{fig:ada}
\end{figure}

Qiao et al.~\cite{qiao2020learning} introduced Meta-learning based Adversarial Domain Augmentation (M-ADA) as an effective single domain generalization method. Recognizing that acquiring data from multiple training domains might not be feasible due to constraints such as budget or privacy concerns, they proposed generating fictitious domains through adversarial training to use in the meta-test phase. As shown in Figure~\ref{fig:ada}, the model consists of a task model and a Wasserstein Autoencoder (WAE). 

The task model with parameters $\theta$ encompasses a feature extractor $F$, which maps the input space to embedding space $z$, and a classifier $C$ that predicts labels from the embedding space $z$, using cross-entropy as its loss function $\mathcal{L}_{task}$. Moreover, $\mathcal{L}_{const}$ imposes a semantic consistency constraint on the adversarially augmented samples $x^{+}$. It ensures that the augmented domains $\mathcal{S}^+$ remain close to the source domain $\mathcal{S}$ within the embedding space. Accordingly, $\mathcal{L}_{const}$ is defined as:
\begin{equation}
    \begin{split}
    \label{eq:l-const}
        \mathcal{L}_{const} = \frac{1}{2} \| z - z^+ \|_2^2 + \infty \cdot 1 \ \{y \neq y^+\},
    \end{split}
\end{equation} 
where $1 \, \{\cdot\}$ is the $0{\text -}1$ indicator function, so $\mathcal{L}_{const}$ will be $\infty$ if $x$ and $x^+$ have different class labels. Essentially, $\mathcal{L}_{const}$ governs the capacity for generalizing beyond the source domain, a measure facilitated by the Wasserstein distance.

In addition, the task model learns from domain augmentations with the assistance of a WAE, denoted as $V$, with parameters $\psi$. $V$ consists of an encoder $Q(x|e)$ and a decoder $G(x|e)$, where $x$ and $e$ denote the input and bottleneck embedding respectively. Also, they utilized Maximum Mean Discrepancy (MMD) as the distance metric ($\mathcal{D}_e$) to minimize the divergence between $Q(x)$ and a priori distribution $P(e)$. Therefore, we can pre-train $V$ using the following optimization on the source domain $\mathcal{S}$:
\begin{equation}
    \begin{split}
    \label{eq:wae}
        \min_{\psi} \ \|G(Q(x)) - x^2\| + \lambda \mathcal{D}_e(Q(x), p(e)),
    \end{split}
\end{equation} 
where $\lambda$ is a hyper-parameter. Following this, the reconstruction error $\mathcal{L}_{relax}$ should be maximized for domain augmentation:
\begin{equation}
    \begin{split}
    \label{eq:l-relax}
        \mathcal{L}_{relax} = \|x^+ - V(x^+)\|^2.
    \end{split}
\end{equation} 

WAEs use the Wasserstein distance metric to measure the distribution distance between the input and reconstruction. As such, the pre-trained V is better equipped to capture the distribution of the source domain, and maximizing $\mathcal{L}_{relax}$ ensures larger domain transportation. Finally, the overall loss function is:
\begin{equation}
    \begin{split}
    \label{eq:l-ada}
        \mathcal{L}_{ADA} = \mathcal{L}_{task}(\theta;x) - \alpha \mathcal{L}_{const}(\theta;z) + \beta \mathcal{L}_{relax}(\psi;x),
    \end{split}
\end{equation} 
where $\alpha$ and $\beta$ are two hyper-parameters used to balance $\mathcal{L}_{const}$ and $\mathcal{L}_{relax}$.

M-ADA offers a novel adversarial domain augmentation approach using relaxed constraints; however, it requires pre-training a Wasserstein autoencoder, which subsequently adds computational overhead.

\subsection{MetaNorm}
While batch normalization is essential in training deep neural networks, small batch sizes and distribution shifts can destabilize batch statistics. MetaNorm, introduced by Du et al.~\cite{du2020metanorm}, ensures effective batch normalization under such conditions. It uses a meta-learning setting to infer adaptive normalization statistics from limited samples instead of relying on direct calculations of batch statistics, which can be unreliable with small batches or distribution shifts. Unlike the transductive batch normalization (TBN) method, which uses the same statistics for both meta-train and meta-test phases, MetaNorm applies a non-transductive approach. Given that test samples may not always be available, MetaNorm is designed to learn to generate statistics only from the support set, and at the meta-test time, it directly applies the model to infer statistics for new tasks.

MetaNorm employs two separate feed-forward networks known as hypernetworks. One is for inferring $\mu$, denoted as $f_{\mu}^{\ell}(\cdot)$, and the other for inferring $\sigma$, denoted as $f_{\sigma}^{\ell}(\cdot)$, in order to find the statistics for each individual channel in each $\ell$ convolutional layer in the meta-learning model. In essence, the hypernetworks accept sample activations as input and produce the estimated statistics as output. During the meta-training phase, the estimated statistics of the support set are applied for normalization of both support and query samples:
\begin{equation}
    \begin{split}
    \label{eq:l-bn}
        a^{\prime} = \gamma \left( \frac{a - \mu_s}{\sqrt{\sigma_s^2 + \epsilon}} \right) + \beta,
    \end{split}
\end{equation} 
where $\gamma$ and $\beta$ are jointly learned with parameters of the hypernetworks during the meta-training and directly applied at meta-test time, and $\epsilon$ is a small scalar to prevent division by zero.

As a final step, it minimizes the following KL term to learn the normalization statistics: 
\begin{equation}
    \begin{split}
    \label{eq:l-kl}
        \sum_i^{|\mathcal{D}^s|} D_{KL}[q_{\phi}(m|a_i) || p_{\theta}(m|\mathcal{D}^s)],
    \end{split}
\end{equation} 
where $q(m|a_i)$ and $p(m|\mathcal{D}^s)$ are defined as a Gaussian distribution based on their respective estimated statistics, and also $a_i$ is a sample from the meta-source domain $\mathcal{D}^s$.
MetaNorm learns to generate its own appropriate statistics and applies them to the samples in the meta-target domain. Notably, in the meta-target domain, the KL term is not used; instead, each example generates its own normalization statistics.

Although adding hypernetworks to infer normalization statistics slightly increases the overall model complexity, MetaNorm overcomes the issues of batch normalization with small batches and domain shifts, offering a promising approach to domain generalization through meta-learning.

\subsection{DADG}
Discriminative Adversarial Domain Generalization (DADG) is a novel framework proposed by Chen et al.~\cite{chen2022discriminative} that integrates Discriminative Adversarial Learning (DAL) for extracting domain-invariant features and Meta-learning-based Cross-Domain Validation (Meta-CDV) for training a robust classifier. DAL aims to learn a domain-invariant feature representation to minimize domain variance. It achieves this by training a feature extractor to confuse a domain discriminator, which is tasked with predicting the domain of the features. Further, as depicted in Figure 1, the Gradient Reversal Layer (GRL) is placed between the feature extractor and the domain discriminator in the DAL component. During forward propagation, the GRL functions as an identity transform, conveying the features unchanged to the discriminator. Conversely, during backpropagation, the GRL reverses the gradient by multiplying it by a negative scalar (e.g., -$\lambda$), thereby updating the parameters of the feature extractor $\theta^m$ into $\theta^{m+1}$. This reversed gradient, originating from the discriminator, encourages the feature extractor to create domain-invariant representations that deceive the discriminator. 

Subsequently, Meta-CDV applies the feature extractor to supervised learning by training a classifier $c_{\varphi}$ with parameters $\varphi$. 
This is done in a meta-learning manner, where the classification model will be trained on the domains seen in the previous step to update $\theta^{m+1}$ to $\theta^{m+2}$ and update $\varphi^{m}$ to $\varphi^{m+1}$. The model's performance is then validated on cross domains to enhance the classification model. It should be noted that this evaluation is performed using the updated parameters $\theta^{m+2}$ and $\varphi^{m+1}$.
In essence, the optimization of the classification model involves the third derivative with respect to $\theta$ and the second derivative with respect to $\varphi$. Particularly, Meta-CDV simulates train/test domain shifts by training the model on seen domains and validating it on a held-out domain, thereby making the classifier more robust.

\begin{figure}[ht]
\centering
\includegraphics[width=0.7\linewidth]{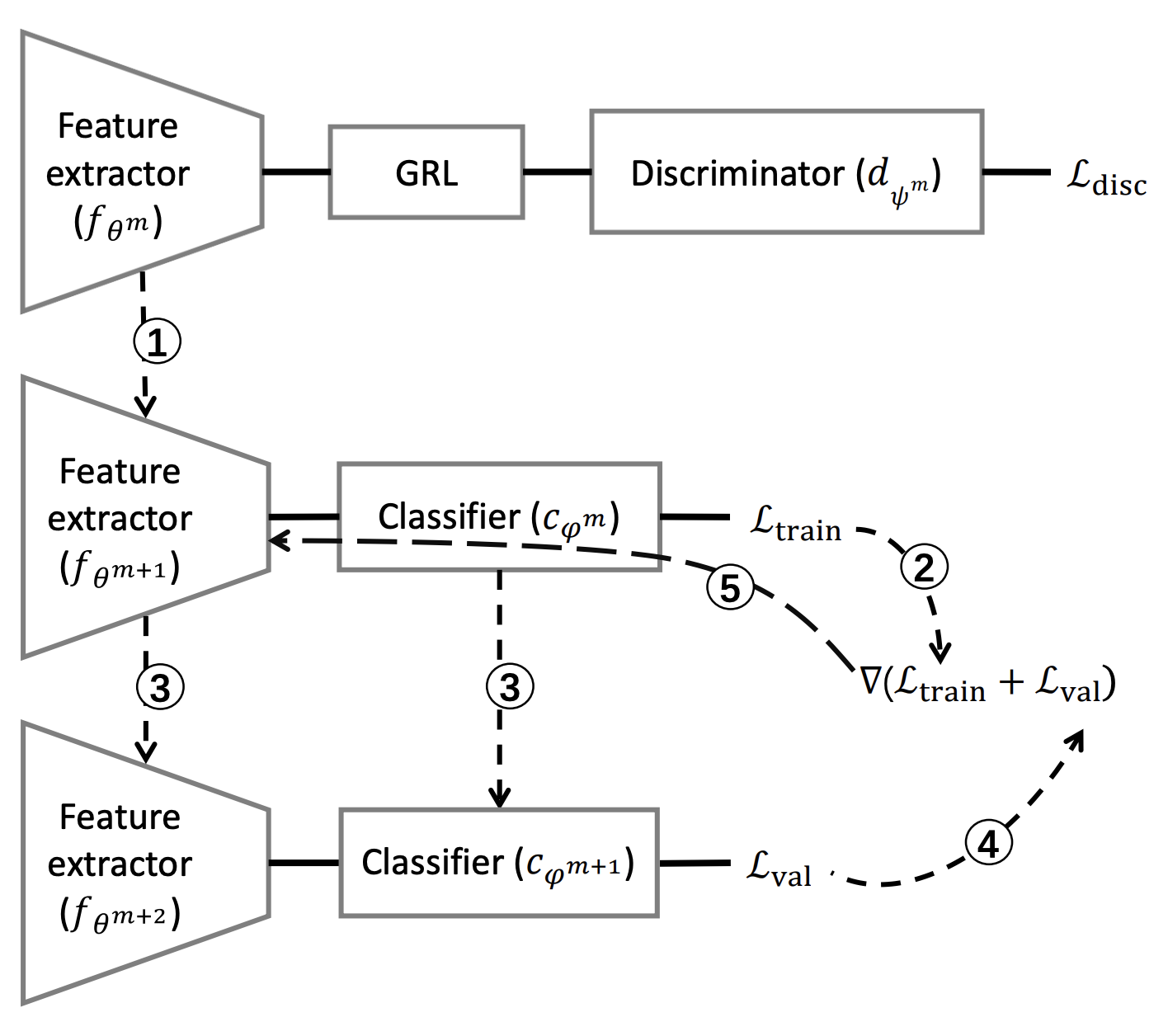}
\caption{Illustration of the DADG framework, which combines Discriminative Adversarial Learning (DAL) featuring a Gradient Reversal Layer (GRL) for creating domain-invariant features with Meta-learning-based Cross-Domain Validation (Meta-CDV) to improve the performance of classifiers across unseen domains.}
\label{fig:dadg}
\end{figure}

DADG learns domain-invariant features using adversarial discriminative learning with GRL, which enables the adversarial mechanism in a simple way. However, the effectiveness of the discriminative adversarial learning component can diminish as the number of source domains increases.

\subsection{Uncertainty-guided Model Generalization}
\begin{figure}[ht]
\centering
\includegraphics[width=0.6\linewidth]{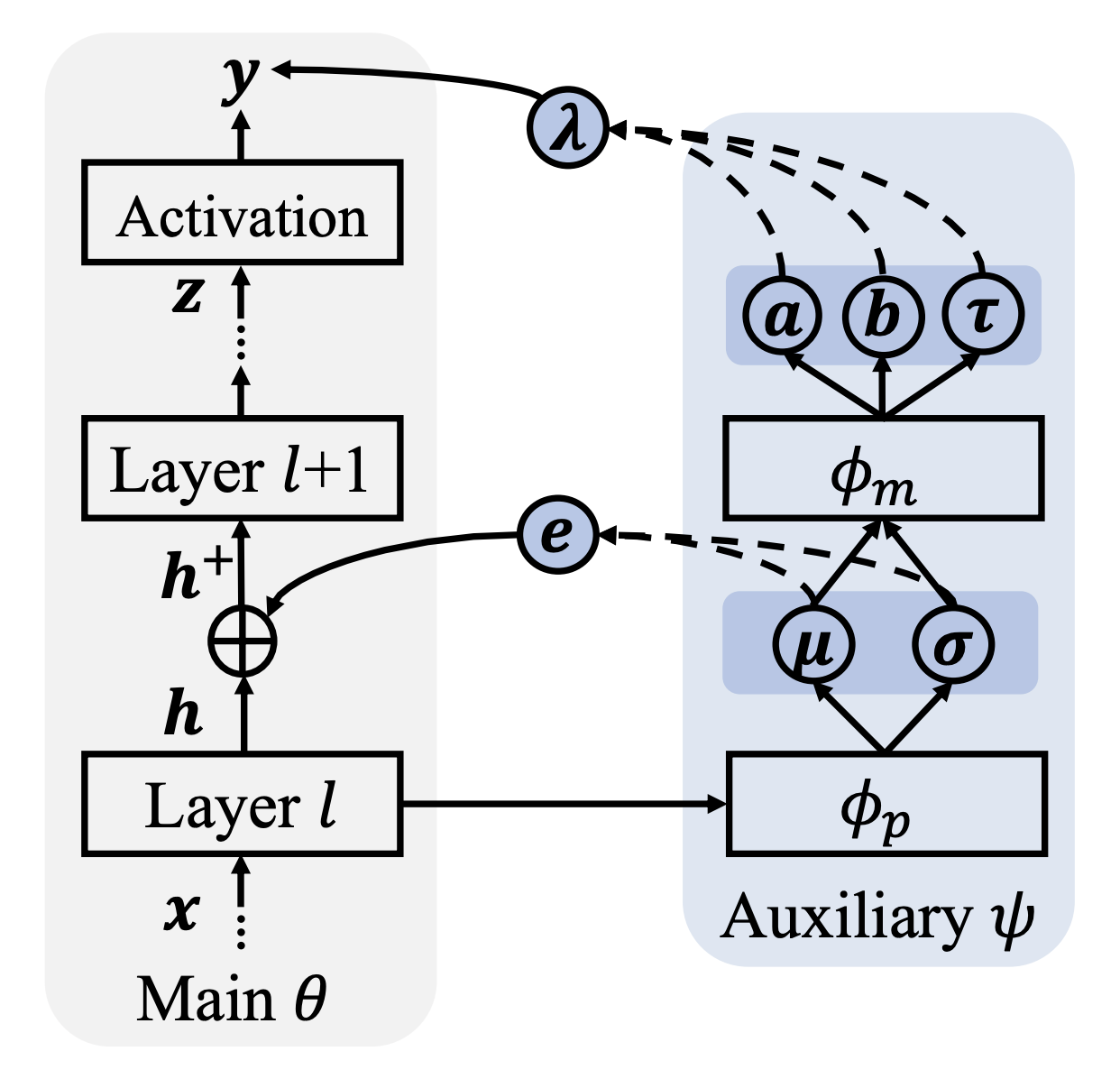}
\caption{Schematic representation of the uncertainty-guided model generalization approach, where an auxiliary network $\psi$ introduces feature perturbations to form a new domain $\mathcal{S}^{+}$. This process utilizes uncertainty in the form of distribution parameters $(\mu, \sigma)$ to guide the creation of domain-augmented data and label mixup, thereby enabling the model to generalize to unseen domains by interpolating between the original and perturbed features and labels.}
\label{fig:uncertainty-guided}
\end{figure}

Qiao et al.~\cite{qiao2021uncertainty} proposed a unique solution for single domain generalization using Bayesian meta-learning, called uncertainty-guided model generalization. The primary strategy of this method is to augment the source domain's capabilities, not only in the input space, as most current data augmentation methods do, but also in the label space, guided by an uncertainty assessment. As illustrated in Figure~\ref{fig:uncertainty-guided}, instead of directly augmenting the input space, they introduce an auxiliary network $\psi = \{\phi_p, \phi_m\}$ to create feature perturbations $h^+$ by adding softplus of Gaussian noise $e \sim \mathcal{N}(\mu, \sigma)$ to generate new domain $\mathcal{S}^{+}$ from domain $\mathcal{S}$. The distribution parameters $(\mu, \sigma)$ represent the uncertainty with respect to the backbone $\theta$. This uncertainty is further utilized to predict learnable parameters $(a, b, \tau)$, which are used to construct learnable label mixup. 
In learnable label mixup, learnable parameters $(a, b)$ are used to define $\lambda \sim {\text{Beta}}(a,b)$ to mixup $\mathcal{S}$ and $\mathcal{S}^{+}$ to achieve in in-between domain interpolations. Notably, learnable parameters $(a, b)$ control the direction and strength of domain interpolations. As a result, the following equations are used:
\begin{equation}
    \begin{split}
    \label{eq:interpolation}
        h^+ &= \lambda h + (1 - \lambda) h^+,\\
        y^+ &= \lambda y + (1 - \lambda) \tilde{y},
    \end{split}
\end{equation} 
where $\tilde{y}$ represents a label-smoothing version of $y$. Particularly, label smoothing is carried out with a probability of $\tau$, which means that we allocate a value of $\rho \in (0,1)$ to the true class and equally distribute $\frac{1 - \rho}{c - 1}$ to the others, where $c$ represents the number of classes. 
Moreover, by learning $a$ and $b$, the model can learn to modulate the Beta distribution to produce optimal $\lambda$ values that result in efficient interpolations between $\mathcal{S}$ and $\mathcal{S}^{+}$.
Accordingly, this process results in consistent domain shifts in both input and output via $\phi_p$ and $\phi_m$, enhancing generalization across unseen domains.

The authors proposed a practical Bayesian meta-learning framework that optimizes $\theta$ and $\psi$ to maximize the posterior $p(\mathcal{S}^+)$. In other words, they utilized Bayesian inference to maximize the posterior of domain augmentations, approximating the distribution of unseen domains.

While uncertainty-guided augmentation provides a mechanism to approximate unseen target distributions, relying solely on randomized sampling procedures to cover the wide range of target domains may be inadequate for real-world distribution shifts.

\subsection{Memory-based Multi-source Meta-learning}
Zhao et al.~\cite{zhao2021learning} proposed a memory-based meta-learning method to address multi-source DG for person re-identification (Re-ID) called Memory-based Multi-source Meta-Learning (${\text M}^{3}{\text L}$). The purpose of person Re-ID is to match persons with the same identity across multiple camera views. Existing works have shown the effectiveness of meta-learning in classification tasks, but its parametric classifier falls short for Re-ID. This is due to the open-set nature of Re-ID tasks, with each domain having numerous and unique identities. Consequently, the ${\text M}^{3}{\text L}$ framework is equipped with a memory-based module that applies the identification loss in a non-parametric way, preventing instability in meta-optimization commonly associated with traditional parametric methods. Specifically, this framework maintains a feature memory $\mathcal{M}^i$ for each source domain $\mathcal{D}^i_\mathcal{S}$ consisting of $n_i$ slots, where each slot saves the feature centroid of the corresponding identity. Identification loss is then computed using the similarities between these features and memory centroids as follows:
\begin{equation}
    \begin{split}
    \label{eq:memory-based}
        \mathcal{L}_M = - \log \frac{\exp{(M[i]^T f(x_i) / \tau)}}{\sum_{k=1}^{n_i} \exp{(M[k]^T f(x_i)/\tau)}},
    \end{split}
\end{equation} 
where $f(\cdot)$ is the feature extractor and $\tau$ is the temperature factor. In addition, the following triplet loss is utilized to train the model:
\begin{equation}
    \begin{split}
    \label{eq:tri}
        \mathcal{L}_{Tri} = [d_p - d_n + \delta]_+,
    \end{split}
\end{equation} 
where $d_p$ represents the Euclidean distance between an anchor feature and a hard positive feature, while $d_n$ signifies the Euclidean distance between an anchor feature and a hard negative feature. Also, $\delta$ is defined as the margin of the triplet loss and $[\cdot]_+ = max(\cdot, 0)$. As a result, the sum of $\mathcal{L}_{Tri}$ and $\mathcal{L}_M$ constitutes the meta-train loss ($\mathcal{L}_{mtr}$) and meta-test loss ($\mathcal{L}_{mte}$).

Furthermore, ${\text M}^{3}{\text L}$ incorporates a meta batch normalization layer (MetaBN) that injects meta-train feature statistics into meta-test features, diversifying them and enabling the model to simulate an expanded range of feature variations. In this way, through iterative generalization processes from meta-train to meta-test domains, the model avoids overfitting due to domain bias and acquires domain-invariant representations that generalize well on unseen domains.

\begin{figure}[ht]
\centering
\includegraphics[width=1\linewidth]{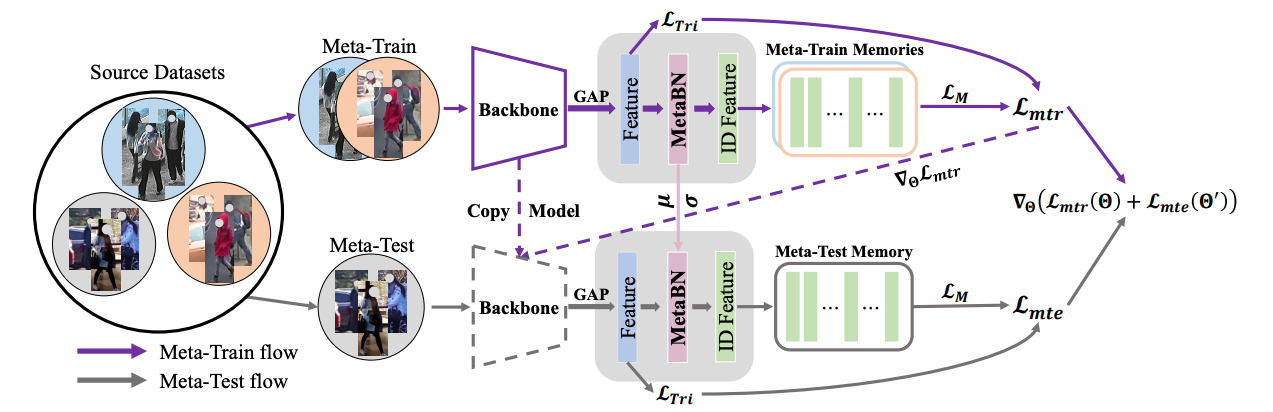}
\caption{Overview of the ${\text M}^{3}{\text L}$ framework designed for person Re-ID. The training process includes dividing the source domains into one meta-test domain and multiple meta-train domains in each iteration. The model utilizes memory-based identification and triplet loss for the meta-training phase and then optimizes the meta-test loss on a copy of the model that has been updated with the meta-train loss (maximization of inter-class distance). MetaBN is also applied during the meta-test stage to enhance feature diversity (maximization of intra-domain distance). Finally, the aggregated meta-train and meta-test losses are used to update the original model for improved domain generalization.}
\label{fig:m3l}
\end{figure}

During training, as shown in Figure~\ref{fig:m3l}, source domains are divided into meta-train and meta-test sets at each iteration, employing a meta-learning approach to simulate the train-test process of DG. Therefore, it is necessary to consider the following optimization to build generalizable representations:
\begin{equation}
    \begin{split}
    \label{eq:m3l}
        \min_{\Theta} \ \mathcal{L}_{mtr}(\Theta) + \mathcal{L}_{mte}(\Theta^{\prime}),
    \end{split}
\end{equation} 
where $\Theta$ represents the network parameters, while $\Theta^{\prime}$ denotes the parameters of the model optimized by $\mathcal{L}_{mtr}$. 

Although MetaBN further enhances generalization ability by diversifying meta-test features, the memory module adds computational overhead in maintaining and updating centroid features for each identity.

\subsection{MetaBIN}
\begin{figure}[h!t]
\centering
\includegraphics[width=0.8\linewidth]{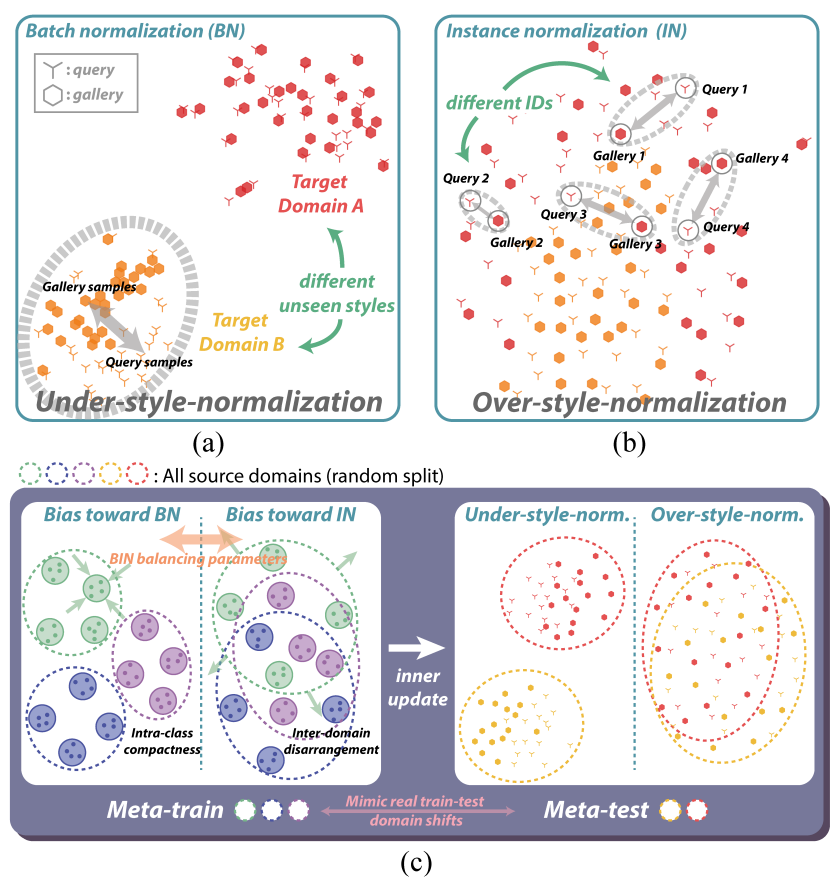}
\caption{Visualized here are scenarios where generalization is unsuccessful, along with the MetaBIN framework. (a) Under-style-normalization where, the BN model fails to recognize identities in novel domains due to a lack of style variation in the training data. (b) Over-style-normalization, where the IN model excessively removes style information, including identity-discriminative features. (c) The main idea behind the MetaBIN framework is to adjust the balance between BN and IN, simulating and learning from both types of normalization errors within a meta-learning framework, thereby enhancing the model's ability to generalize to unseen domains without overfitting to source domain styles.}
\label{fig:bnin}
\end{figure}

Choi et al.~\cite{choi2021meta} proposed Meta Batch-Instance Normalization (MetaBIN) as an effective heterogeneous single domain generalization method for person re-identification (Re-ID). The main goal is to simulate unsuccessful generalization scenarios by combining batch-instance normalization layers with meta-learning to address challenging cases caused by both batch and instance normalization layers. 
A key feature of MetaBIN is its learnable balancing parameters between Batch Normalization (BN) and Instance Normalization (IN), which, depending on their bias, cause the DG model to experience under-style-normalization and over-style-normalization scenarios during meta-learning. Under-style-normalization occurs in the BN model when the model struggles to distinguish identities of samples with unexpected styles from unseen target domains. Conversely, over-style-normalization arises in the IN model, which, while efficient at eliminating instance-specific style information, can inadvertently filter out discriminative information; refer to Figure~\ref{fig:bnin} for more information.

MetaBIN consists of a classifier $g_{\phi}$ with parameter $\phi$ to predict identities and a feature extractor $f_{\theta}$ with parameters $\theta$ where $\theta = (\theta_f, \theta_{\rho})$. Here, $\theta_{\rho}$ signifies the balancing parameters between BN and IN, and $\theta_f$ represents the remaining parameters.

In the base model updating process, classifier parameters $\phi$ and feature extractor parameters $\theta_f$ are updated by minimizing the cross-entropy loss along with the triplet loss according to the following:
\begin{equation}
    \begin{split}
    \label{eq:triplet}
        \mathcal{L}_{tri} = [d_p - d_n + \delta]_+,
    \end{split}
\end{equation} 
where $d_p$ represents the Euclidean distance between an anchor feature and the hardest positive sample, while $d_n$ signifies the Euclidean distance between an anchor feature and the hardest negative sample. Also, $\delta$ is defined as the margin of the triplet loss and $[\cdot]_+ = max(\cdot, 0)$.

In the meta-learning step, source domains are split into meta-train set $\mathcal{D}_{mtr}$ and meta-test set $\mathcal{D}_{mte}$. First, in the meta-training phase, only $\theta_{\rho}$ is updated to obtain $\theta^{\prime}_{\rho}$ using $\mathcal{D}_{mtr}$ to simulate over-style-normalization and under-style-normalization cases via the aggregation of the following losses: the intra-domain scatter loss ($\mathcal{L}_{scat}$), the inter-domain shuffle loss ($\mathcal{L}_{shuf}$), and the triplet loss ($\mathcal{L}_{tri}$). The intra-domain scatter loss is used to spread the feature distribution for each domain by minimizing the cosine similarity between each sample and its domain centroid. The inter-domain shuffle loss is introduced to shuffle or mix up the distributions across different source domains by minimizing the distance between anchor samples and inter-domain negatives while maximizing the distance between anchor samples and intra-domain negatives. In summary, the intra-domain scatter loss $\mathcal{L}_{scat}$ and inter-domain shuffle loss $\mathcal{L}_{shuf}$ are used to simulate over-style-normalization scenarios where styles are confused. At the same time, adding the triplet loss enhances intra-class compactness regardless of style differences to simulate under-style-normalization. Next, in the meta-testing phase, the model with parameters $\theta_f$ and $\theta^{\prime}_{\rho}$ is evaluated on $\mathcal{D}_{mte}$, and $\theta_{\rho}$ will be updated using $\mathcal{L}_{tri}$. Note that $\theta_{\rho}$ contains channel-wise balancing parameters $\rho$ for each normalization layer, with some channels biasing $\theta_{\rho}$ toward IN to remove unessential style info, while others retain BN properties.

MetaBIN improves generalization to unseen domains without needing extra networks or data augmentation; however, it may not perform well on datasets with complex variations or highly different styles between domains, as it relies on the simulated scenarios to learn a robust representation.

\vspace{7pt}
\section{Datasets and Evaluations}
\label{sec:evaluation}
In the DG and domain adaptation (DA) fields, datasets are specifically constructed to represent various 'domains'—each a unique distribution of data that captures a certain variation in the input space. These variations can include changes in visual appearance, sensor modality, environmental conditions, or even different tasks. The purpose of these multi-domain datasets is to simulate real-world scenarios where a model trained on limited domains must perform well on another unseen domain. The richness and diversity of domains within these datasets are pivotal to developing and benchmarking algorithms that can generalize beyond their training data.

Several important datasets, commonly used in the fields of DG and DA, have been examined in studies that leverage meta-learning for DG. Given the fundamental overlap between DA and DG, datasets from DA can also be applied to DG purposes, and vice versa. Each dataset, associated with a specific application, is summarized in Table~\ref{table:dataset} and will be discussed in this section. This discussion will be followed by an overview of the evaluation strategies commonly employed in DG, which differ from the more straightforward evaluation systems used in DA.

\begin{table*}[ht]
\renewcommand{\arraystretch}{1.3}
\caption{Overview of Key Datasets Used in Domain Generalization Based on Their Application, Highlighting the Number of Domains, Number of Classes, Number of Samples, and Their Description.}
\label{table:dataset}
\centering
\resizebox{1\textwidth}{!}{
\begin{tabular}{c c c c c}
\hline
 & \# Domains & \# Classes & \# Samples & Description \\ 
\hline
\hline
\textbf{Handwritten Digit Recognition} & & & & \\
\hline
Rotated MNIST~\cite{ghifary2015domain}  & 6 & 10 & 70,000 & Rotations (0, 15, 30, 45, 60, and 75) \\
Digits-Five~\cite{zhou2020learning}  & 5 & 10 & 215,695 & MNIST~\cite{lecun1998gradient}, MNIST-M~\cite{ganin2015unsupervised}, SVHN~\cite{netzer2011reading}, SYN~\cite{ganin2015unsupervised}, and USPS~\cite{hull1994database} \\
\hline
\hline
\textbf{Object Recognition} & & & & \\
\hline
VLCS~\cite{fang2013unbiased}  & 4 & 5 & 10,729 & VOC2007~\cite{everingham2010pascal}, LabelMe~\cite{russell2008labelme}, Caltech-101~\cite{fei2004learning}, and SUN09~\cite{xiao2010sun} \\
Office-Home~\cite{fang2013unbiased}  & 4 & 65 & 15,588 & Art, Clipart, Product, and Real-World \\
PACS~\cite{li2017deeper}  & 4 & 7 & 9,991 & Photo, Art painting, Cartoon, and Sketch \\
CIFAR-10-C~\cite{hendrycks2019benchmarking}  & - & 10 & 60,000 & Artificial Corruptions of CIFAR-10~\cite{krizhevsky2009learning} \\
Visual Decathlon~\cite{rebuffi2017learning}  & 10 & - & 1,659,142 & 10 Different Datasets (\cite{maji2013fine, krizhevsky2009learning, munder2006experimental, cimpoi2014describing, stallkamp2012man, russakovsky2015imagenet, lake2015human, netzer2011reading, soomro2012ucf101, bilen2016dynamic, nilsback2008automated}) \\
\hline
\hline
\textbf{Action Recognition} & & & & \\ 
\hline
IXMAS~\cite{weinland2006free}  & 5 & 11 & 1,650 &  5 Different Camera Views, and 11 Different Subjects (\cite{li2018domain}) \\
\hline
\hline
\textbf{Semantic Segmentation} & & & & \\ 
\hline
SYNTHIA~\cite{ros2016synthia}  & 15 & 13 & 2,700 &  5 Different Locations, and 5 Different Weather Conditions (\cite{volpi2018generalizing}) \\
\hline
\hline
\textbf{Person Re-Identification} & & & & \\ 
\hline
Market-Duke~\cite{ristani2016performance}  & 2 & - & 69,079 &  Market-1501~\cite{zheng2015scalable} and DukeMTMC-ReID~\cite{zheng2017unlabeled} \\
\hline
\hline
\textbf{Sentiment Classification} & & & & \\ 
\hline
Amazon Reviews~\cite{blitzer2006domain}  & 4 & 2 & $>$ 340,000 &  Books, DVDs, Electronics, and Kitchen Appliances \\
\hline
\end{tabular}
}
\end{table*}

\subsection{Datasets}
In the DG landscape, datasets serve as critical benchmarks for evaluating algorithm performance across varied scenarios. We begin with Rotated MNIST~\cite{ghifary2015domain}, a synthetic dataset based on the original MNIST. It comprises six domains, each containing images of various digits modified by rotations ranging from 0 to 90 degrees at 15-degree intervals. This dataset is instrumental in assessing DG algorithms for handwritten digit recognition. Additionally, to serve a similar purpose, another dataset, known as Digits-Five~\cite{zhou2020learning}, has been introduced, which includes five sub-datasets: MNIST~\cite{lecun1998gradient}, MNIST-M~\cite{ganin2015unsupervised}, SVHN~\cite{netzer2011reading}, SYN~\cite{ganin2015unsupervised}, and USPS~\cite{hull1994database}, where each of them can be considered a different domain.

\begin{figure}[ht]
\centering
\includegraphics[width=1\linewidth]{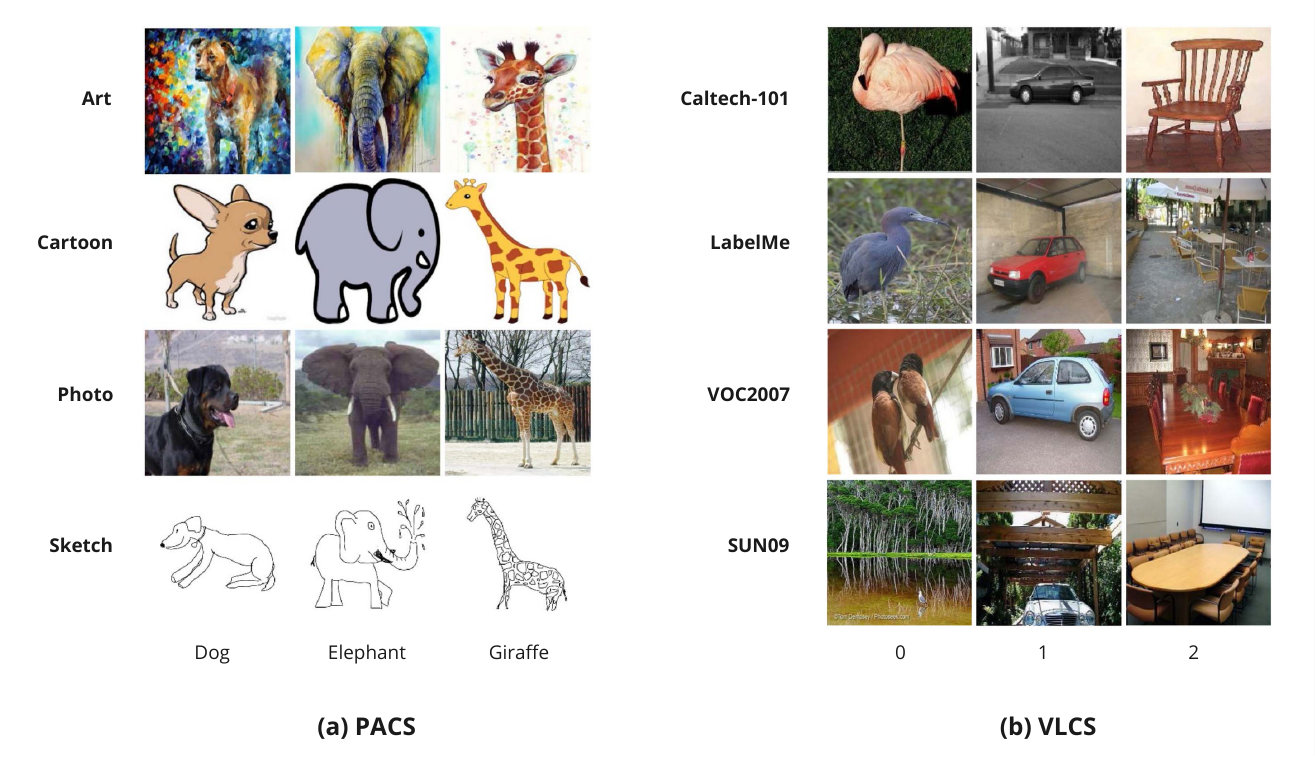}
\caption{Example images from two prominent domain generalization benchmarks, illustrating different types of domain shifts. In (a), the PACS dataset highlights domain shifts primarily due to changes in image style. In (b), the VLCS dataset reveals domain shifts caused by variations in environment, scene, and viewpoint, reflecting dataset-specific biases.}
\label{fig:datasets}
\end{figure}

For object recognition, a task particularly sensitive to domain shifts, notable benchmarks include VLCS~\cite{fang2013unbiased}, Office-Home~\cite{fang2013unbiased}, PACS~\cite{li2017deeper}, CIFAR-10-C~\cite{hendrycks2019benchmarking}, and VD~\cite{rebuffi2017learning}. VLCS contains images from four domains (VOC2007~\cite{everingham2010pascal}, LabelMe~\cite{russell2008labelme}, Caltech-101~\cite{fei2004learning}, and SUN09~\cite{xiao2010sun}) (see Figure~\ref{fig:datasets}) and five classes. Office-Home includes images of objects from Art, Clipart, Product, and Real-World domains, spanning 65 categories. PACS covers four contrasting domains (Photo, Art painting, Cartoon, and Sketch as depicted in Figure~\ref{fig:datasets}) with seven object classes. VD contains ten diverse domains, including handwritten characters, pedestrians, traffic signs, etc., with varying image categories and sizes suitable for heterogeneous DG. Also, CIFAR-10-C~\cite{hendrycks2019benchmarking} is a robustness benchmark comprising CIFAR-10~\cite{krizhevsky2009learning} test images corrupted through 19 distortion types across five severity levels. 

IXMAS~\cite{weinland2006free} dataset addresses the domain generalization challenge in action recognition, which relies on learning generalizable representations within video understanding. It includes recordings of 11 actions performed by various actors, captured from multiple angles and with different cameras, to introduce domain shifts.

In autonomous vehicle development, semantic segmentation is essential, yet deep neural networks have not fully bridged the performance gap in unexplored environments. SYNTHIA~\cite{ros2016synthia} dataset, with its synthetic images of different locations and weather conditions, supports research to overcome this limitation.

Person Re-Identification (Re-ID) is another area where DG is paramount, especially for security and surveillance. While traditional Re-ID has been confined to the same camera views, cross-dataset Re-ID seeks to generalize from source to target views across varied resolutions, viewpoints, and lighting conditions. Key datasets like Market-1501~\cite{zheng2015scalable} and DukeMTMC-ReID~\cite{zheng2017unlabeled} are essential in the advancement of this field.

Lastly, the Amazon Reviews~\cite{blitzer2006domain} dataset, containing reviews from domains such as books, DVDs, electronics, and kitchen appliances, is crucial for sentiment classification in natural language processing.

\subsection{Evaluations}
For evaluating domain generalization algorithms, three principal strategies are commonly employed: (i) \textit{Leave-one-domain-out validation}, which is the most prevalent strategy that is applicable when multiple source domains are available during training. It involves leaving out one source domain as the validation set while using the others for training. However, the final results heavily rely on the selection of the validation domain, leading to results that may lack stability. (ii) \textit{Training-domain validation}, where a subset of the training data is held out for validation to select the best model. However, since there is still a divergence between real-world unseen data and the training subset, it may not achieve optimal performance. (iii) \textit{Test-domain validation}: which involves using a random subset of the target domain data for model selection. This method can lead to the best performance since the validation and test data share the same distribution. However, it presupposes access to the target domain data, which may not be feasible in real-world applications. Using test domain data for model selection can yield misleading results, as an unrepresentative or excessively easy subset may result in optimistic performance, or a challenging subset may lead to pessimistic results~\cite{gulrajani2020search}. Careful interpretation of results obtained using this method is therefore necessary. 

In general, the average accuracy is used as the primary metric to report the performance of a DG model across held-out domains. For image classification on Cifar-10-C~\cite{hendrycks2019benchmarking}, the mean Corruption Error (mCE) metric determines model robustness to corruptions, while Relative mCE (RmCE) compares corruption robustness relative to a baseline by accounting for clean data performance. This enables fairer comparisons between models. For semantic segmentation models evaluated on the SYNTHIA~\cite{ros2016synthia} dataset, the standard mean Intersection Over Union (mIoU) is calculated for each unseen domain to quantify segmentation accuracy. Additionally, the VD-score metric assesses models trained on the diverse VD~\cite{rebuffi2017learning} dataset by evaluating performance across its ten distinct image classification tasks.

\vspace{7pt}
\section{Applications}
\label{sec:applications}
In real-world scenarios, finding sufficient labeled data to effectively train a model often proves challenging. This difficulty is compounded when the model encounters OOD data during testing. As such, domain generalization through meta-learning has emerged as a highly desirable solution. It enables a model to be trained on one or multiple domains and to be applied to unseen domains without the need for retraining. This approach not only alleviates the risk of overfitting but also diminishes the costs and time associated with domain-specific retraining, thus boosting the model's adaptability and operational efficiency.

The applicability of domain generalization models is evident across various disciplines, such as medical imaging, intelligent fault diagnosis, computer vision, and natural language processing. For example, Liu et al.~\cite{liu2020shape} utilized meta-learning for domain generalization to segment prostate MRI images, demonstrating its potential in medical analysis. Ren et al.~\cite{ren2023meta} proposed Meta-GENE, a model-agnostic meta-learning framework designed for fault diagnosis in industrial prognostics and health management scenarios, which can operate in both homogeneous and heterogeneous DG settings.
Wang et al.~\cite{wang2020meta} applied the technique for cross-lingual semantic parsing, and Balaji et al.\cite{balaji2018metareg} utilized it for sentiment analysis within the Amazon Review dataset. In the area of person re-identification (Re-ID), both Choi et al.~\cite{choi2021meta} and Zhao et al.~\cite{zhao2021learning} leveraged meta-learning for DG to recognize individuals across varying postures and perspectives. Moreover, meta-learning for DG has been applied in cross-view action recognition by Li et al.~\cite{li2019episodic} and robot control through reinforcement learning by Li et al.~\cite{li2018learning}. Qiao et al.~\cite{qiao2020learning} showcased its utility in semantic segmentation for street scenes, and Dou et al.~\cite{dou2019domain} explored its application in multi-site brain tissue segmentation.

Further extending the applications, DG has been employed in face anti-spoofing~\cite{shao2019multi}, image compression~\cite{zhang2021out}, and various medical diagnostics, including Parkinson's disease detection~\cite{muandet2013domain}, activity recognition~\cite{erfani2016robust}, chest X-ray analysis~\cite{mahajan2021domain}, and EEG-based seizure detection~\cite{ayodele2020supervised}. Other studies have used domain generalization for speech utterance recognition~\cite{shankar2018generalizing, piratla2020efficient}, fault diagnosis~\cite{li2020domain, zheng2020deep, liao2020deep}, and brain-computer interaction~\cite{han2021domain}. Additionally, DG has found its place in brain-computer interfacing~\cite{han2021domain}, and promising advancements have been made in time series forecasting for financial markets~\cite{du2021adarnn}.

The extensive applications of meta-learning for DG position it as a focal point of interest for future research and commercial endeavors. The algorithms developed under this paradigm equip models with rapid generalization and adaptation capabilities, even in the absence of target domain data. Such a capacity is a stepping stone towards creating more powerful AI systems capable of tackling diverse real-world problems. By integrating domain generalization with meta-learning, we can unlock zero-shot learning capabilities for novel, unseen domains, enabling generalization across various tasks and domains without requiring additional data.

\vspace{7pt}
\section{Discussion and Future Directions}
\label{sec:discussion}
This survey has examined diverse strategies for DG using meta-learning. The presented taxonomy provides a structured view of the field, categorizing approaches based on the treatment of the input domain and classifier training strategies, as discussed in Section~\ref{sec:taxonomy}. 
Initially, DG models employing meta-learning predominantly focused on the minimization of inter-domain distances and the minimization of intra-class distances, aiming to find a common representation across domains and cohesive grouping of class instances, as this approach is the most intuitive way to develop DG methods using meta-learning. 
However, recent algorithms have been exploring the effectiveness of the Maximization of Inter-Class Distances or the Maximization of Intra-domain Distances to fully exploit the advantages of meta-learning. 
For instance, MASF~\cite{dou2019domain} leverages the Maximization of Inter-Class Distances. It is the first model in meta-learning for DG that follows this paradigm, resulting in improved class discriminability and enhanced generalization to unseen domains.
Furthermore, M-ADA first introduced the concept of maximizing intra-domain distances through meta-learning, laying the groundwork for future models that employ this strategy. This approach aims to diversify the input feature space, thereby enhancing the robustness of the feature extractor, particularly when there are limited training domains available. Recently, methods such as Memory-Based Multi-Source Meta-Learning ($M^{3}L$)~\cite{zhao2021learning} and MetaBIN~\cite{choi2021meta} have been developed, leveraging the Maximization of both Intra-Domain and Inter-Class Distances. This strategic combination is designed to diversify the feature space within each domain while simultaneously ensuring that features remain discriminative for classification tasks. These methods are beneficial for use cases where different classes are not easily distinguishable, such as Re-Identification (Re-ID), by further diversifying input features and enhancing class discriminability.

The choice of strategy in DG models tends to vary depending on the nature and number of available training domains. When multiple diverse domains are available for training, employing a technique that minimizes inter-domain distances is a natural choice, as it exploits the diversity of the available data to extract domain-invariant features. 
On the other hand, if the training data is limited to a few domains, models that maximize inter-domain distances are preferable, as they focus on augmenting the input or feature space to create a generalizable model capable of performing well across various domains during inference.
In particular, recent advances in generative AI allows more realistic synthetic data generation and diversification, which further enhance algorithms that maximize intra-domain distances.  

Additionally, in scenarios where data points within a dataset are highly similar to each other, it is beneficial to utilize models that emphasize the Maximization of Inter-Class Distances. This approach encourages a clear separation between classes, enabling the model to learn effective embedding representations that facilitate discrimination between classes based on distance.

To alleviate the weaknesses of existing meta-learning approaches for DG which has been explored in Figure~\ref{fig:decision}, a promising future direction is to create a novel distance metric that that captures the intra- and inter-domain characteristics. This metric can be utilized as a guidance for selecting appropriate methods to enable rapid adaptation to new domains. 

It is also worth mentioning that the exploration of biologically plausible models in meta-learning has gained some attention and aims to bridge the gap between artificial learning systems and the mechanisms underlying learning in biological organisms. While artificial neural networks have achieved impressive results on meta-learning benchmarks, they differ substantially from the learning processes in biological systems, such as the human brain. As of recent, there has been growing interest in developing meta-learning models that are more biologically grounded and mimic the human learning system.
One such model incorporates a memory system designed to prevent catastrophic forgetting, functioning similarly to an episodic memory system~\cite{khoee2023meta}. This model enhances the generalization capabilities of spiking neural networks by simulating an efficient episodic memory capable of storing vast amounts of information and linking similar underlying patterns.
A promising direction for future development is to create innovative policies for memory adaptation within this model to handle more complex tasks, such as DG. By implementing these policies, the model can better ignore noise and outlier data and map similar classes from different domains to the same memory representation. This would empower the model to effectively tackle data from diverse domains, enhancing its generalization capabilities.

DG is a critical challenge in federated learning, where multiple clients collaboratively train a model without sharing their raw data, enhancing privacy and security~\cite{zhang2021survey}. However, the domain of the target test data on the server can differ greatly from the training data of each client. This discrepancy leads to decreased performance of the federated model on the target domain. RFDG~\cite{guan2023rfdg} has integrated domain generalization into federated learning via a reinforcement learning-based feature decorrelation policy that enables clients to reweight samples from a global perspective to learn domain-invariant knowledge. The replay mechanism also addresses challenges posed by mini-batch training. This field represents a potential area where we should incorporate meta-learning for DG to enhance adaptability to different clients with limited labeled data. By leveraging meta-learning techniques, the federated model can quickly adapt to new clients by learning from a small number of labeled examples, even when there are significant distributional shifts between the source and target domains. Accordingly, meta-learning can be used to learn a shared feature extractor that is robust to domain shifts, allowing the model to extract transferable knowledge from the source domains and effectively apply it to the target domain, thereby improving its ability to generalize to diverse and unseen domains encountered in real-world federated learning scenarios.

Although the presented approaches have made notable progress in domain generalization using meta-learning, there are still interesting avenues for future research. One such direction is Generalizable Label Distribution Learning (GLDL), which aims to learn a Label Distribution Learning (LDL) model that can generalize well to unseen target domains. LDL is a machine learning paradigm that assigns a label distribution to each instance, indicating the relative importance or description degree of each label. By reflecting the varying degrees of association between labels and instances, LDL provides a more detailed representation of label information. Consequently, leveraging the full distribution of labels can enhance model performance on complex tasks. A recent work by Zhao et al.~\cite{zhao2023generalizable} has expanded the scope of domain generalization research by addressing the specific challenges of LDL. In this work, the DICE framework was proposed, which learns to extract domain-invariant feature-label correlations and label-label correlations. It achieves this by aligning the prior distributions and label correlation matrices across different source domains. This work opens up new possibilities for applying meta-learning techniques to the GLDL problem, potentially enabling more robust and generalizable LDL models that can handle distribution shifts between source and target domains. Exploring the integration of meta-learning strategies with GLDL could lead to further advancements in this emerging area of research.

It is also important to note another key challenge that still needs to be addressed in this area of research: enabling algorithms to go beyond pattern recognition to understand and exploit data causality~\cite{rothenhausler2021anchor}. This understanding is essential for models to make consistent predictions across domains, particularly when existing correlations are unreliable~\cite{sheth2022domain, lv2022causality, chen2023meta}. 
Therefore, it is essential to provide meta-learning models with proper inductive biases that can guide the learning process toward causal inference. Inductive biases help in shaping the hypothesis space, enabling the model to prioritize learning generalizable patterns that reflect underlying causal relationships as opposed to spurious correlations. 
Despite challenges, meta-learning has valuable potential for DG if steered to grasp causal structures rather than merely mimic data. Developing meta-learning frameworks that inherently capture causal relationships will be pivotal for achieving true domain-agnostic capabilities.

\vspace{7pt}
\section{Conclusions}
\label{sec:conclusion}
Meta-learning has experienced rapid growth in interest in recent years and has been applied to many significant research topics in machine learning, including DG. In this paper, we focus on meta-learning for DG, a promising field that has attracted many researchers. We provide a comprehensive review of existing methods and present a detailed taxonomy based on two primary axes crucial for designing effective models, as well as an overview of relevant datasets, benchmarks, and applications.
The taxonomy provided in this survey offers a roadmap for understanding the diverse strategies that underpin current research in the field. It is clear that both the feature extraction and classifier training processes are critical to the development of DG models.
Furthermore, we share insights from our analysis of these methods and identify potential research challenges that could guide future research directions. We hope that this survey will assist newcomers and practitioners in navigating this growing field and will also highlight opportunities for future research.

\bibliography{sn-bibliography}

\end{document}